%% file: arxiv.tex
\documentclass{article}
\usepackage{iclr2027_conference,times}
\iclrfinalcopy
\usepackage[utf8]{inputenc} 
\usepackage[T1]{fontenc}    
\usepackage{hyperref}       
\usepackage{url}            
\usepackage{booktabs}       
\usepackage{amsfonts}       
\usepackage{nicefrac}       
\usepackage{microtype}      
\usepackage{xcolor}         
\usepackage[utf8]{inputenc}
\usepackage{booktabs}   
\usepackage{tabularx}   
\usepackage{graphicx}
\usepackage{multirow}
\usepackage{graphbox} 
\usepackage{amsmath} 
\usepackage{amssymb}
\usepackage{amsthm}
\usepackage{booktabs}
\usepackage[table]{xcolor}
\usepackage{siunitx}
\usepackage{xspace}
\usepackage{float}
\newcommand{\model}{\textsc{MUSE}\xspace}

\usepackage{algorithm}
\usepackage{algpseudocode}
\usepackage{amsmath,amssymb}
\usepackage{xcolor}
\usepackage{float}

\floatstyle{ruled}
\restylefloat{algorithm}

\DeclareMathOperator{\Gap}{Gap}
\DeclareMathOperator{\TopThree}{Top3}
\DeclareMathOperator{\ValidName}{ValidName}
\newtheorem{definition}{Definition}
\usepackage{tcolorbox}
\tcbuselibrary{listings, breakable, skins}

\newtcblisting{promptbox}[1][]{
  colback=gray!5!white,       
  colframe=gray!75!black,     
  fonttitle=\bfseries,        
  title={#1},                 
  breakable,                  
  enhanced,                   
  listing only,               
  listing options={
    basicstyle=\ttfamily\small, 
    breaklines=true,            
    breakatwhitespace=true,     
    columns=fullflexible,       
    keepspaces=true,            
    showstringspaces=false      
  }
}
\title{\model: Benchmarking Assembly Structure and Interface Geometry in Multi-Component Text-to-CAD Generation
}

\author{
Xiaoyu Dong$^{1}$ \quad
Zeng Chenbo$^{2}$ \quad
Zhi Li$^{3}$ \quad
Xiao-Ming Wu$^{1,*}$ \\
$^{1}$The Hong Kong Polytechnic University \quad
$^{2}$Lingnan University \quad
$^{3}$Curvature Flow Co., Ltd.\\
\textsuperscript{*}Corresponding authors
}
\begin{document}

\maketitle


\begin{abstract}


Large language models (LLMs) have recently advanced text-driven
3D generation, yet generating CAD models that satisfy practical
design requirements remains challenging. Existing benchmarks
focus primarily on single-part CAD models and geometric
similarity, providing limited assessment of assembly structure
and interface geometry. To address this gap, we introduce
\model, a Text-to-CAD benchmark for generating multi-component,
editable boundary representation (B-Rep) assemblies.
\model pairs practical design instances with structured
\textit{Design Specifications} and evaluates generated outputs
across four dimensions: code executability, geometric validity,
structural and interface fidelity, and functional consistency.
We further introduce Structure and Interface Evaluation (SIE),
which combines reference Physical Assembly Graphs with direct
B-Rep analysis to assess component correspondence, connectivity,
interface types, and interface clearances. Experiments on proprietary
and open-weight LLMs show that high execution success does not
necessarily translate into correct assembly structure or
interface geometry. Together, \model provides a benchmark and
evaluation framework for studying multi-component Text-to-CAD
generation beyond shape similarity. 
Our project website, including the leaderboard, dataset, and code, is available at \url{https://dong7313.github.io/muse-benchmark/}.


\end{abstract}


\input{chapter/1._Introduction}
\input{chapter/2.Related_Work}

\input{chapter/3.Benchmark}

\input{chapter/4.Experiments}

\input{chapter/5.Conclusion}

\subsection*{AI use statement}

We used generative AI tools for synthetic dataset generation
and benchmark evaluation.
Claude Opus 4.7 assisted in augmenting expert-written
CadQuery scripts to create design variants.
An LLM analyzed STEP models to generate draft Physical
Assembly Graphs, and GPT-5.5 generated natural-language
design descriptions.
Human experts reviewed and corrected both the descriptions
and the graphs comprising the final Design Specifications.
We also used a VLM to evaluate functional consistency
from standardized engineering drawings, as described
in our evaluation protocol.
Additionally, generative AI assisted with manuscript editing,
terminology standardization, and LaTeX formatting.

\subsection*{Ethics statement}

This work develops a benchmark for evaluating multi-component
CAD generation.
Its evaluation results should not be interpreted as
certification of manufacturing safety or physical assembly
feasibility.
In particular, agreement in assembly structure and interface
geometry does not establish load-bearing capacity, material
suitability, or safe performance in real-world use.
Generated designs require application-specific engineering
validation before fabrication or deployment.
LLM-generated annotations and VLM-based assessments may
also contain errors; we describe expert review of the
Design Specifications and evaluate VLM--human agreement
to characterize these limitations.

\subsection*{Reproducibility statement}

The dataset construction process is described in
Section~\ref{sec:dataset_construction}, with prompting
examples provided in Appendix~\ref{app:prompt_templates}.
Section~\ref{sec:evaluation} defines the evaluation protocol
and metrics, while Appendix~\ref{app:evaluation_details}
provides detailed notation and implementation procedures.
Together, these descriptions document the construction
of Design Specifications and the evaluation of generated
CAD models, including assembly structure and interface
geometry.


\bibliographystyle{iclr2027_conference}
\bibliography{chapter/ref}

\appendix
\input{chapter/Appendix}


\end{document}

%% file: chapter/1._Introduction.tex
\begin{figure}[htbp]
  \centering
  \includegraphics[width=\linewidth]{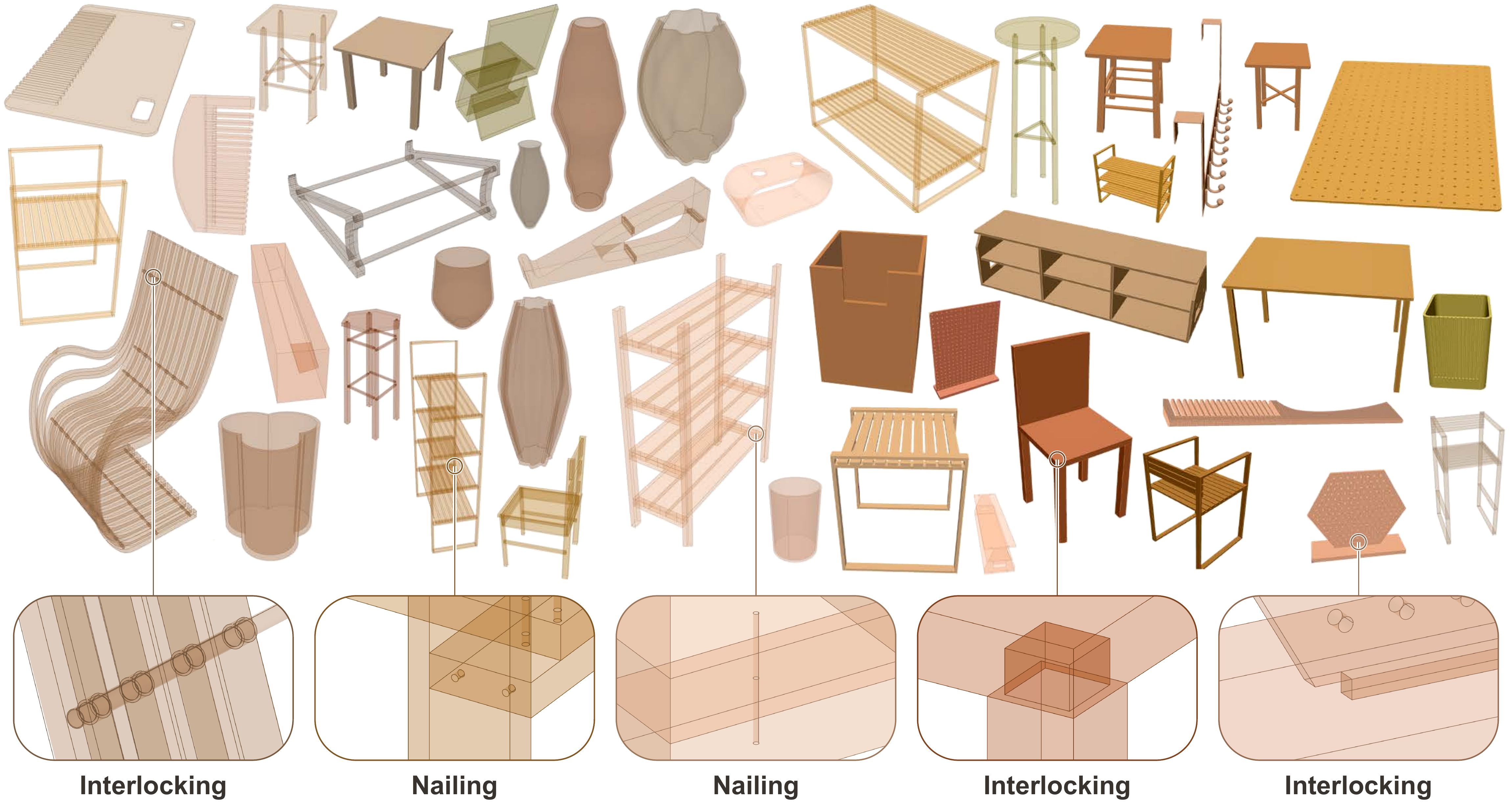}
  \caption{Examples of multi-component designs in \model, illustrating explicit assembly structure and interface geometry.}
\end{figure}

\section{Introduction}
\label{sec:intro}

Text-to-CAD, the task of generating computer-aided design (CAD)
models from natural-language descriptions, is emerging as a
promising direction for AI-assisted 3D modeling~\cite{text2cad}.
Beyond visual plausibility, Text-to-CAD aims to produce
\textbf{design instances} grounded in practical use and
engineering requirements.
For multi-component designs, valid individual geometries
alone are insufficient: the components must also connect
and fit together as intended.
This shifts the focus toward engineering usability, requiring
the generated model to capture the intended shape while
satisfying usage-level constraints.

Although Text-to-CAD datasets~\cite{text2cad} have grown
substantially in recent years, many still focus on single-part
CAD models, providing limited coverage of the interdependent
constraints in practical multi-component designs.
Such designs require not only valid part geometry, but also
correct connectivity, interface types, and interface clearances.
Constructing design instances that explicitly capture these
requirements demands substantial input from experienced
designers, making them costly to curate.

\begin{figure}[htbp]
  \centering
  \includegraphics[width=\linewidth]{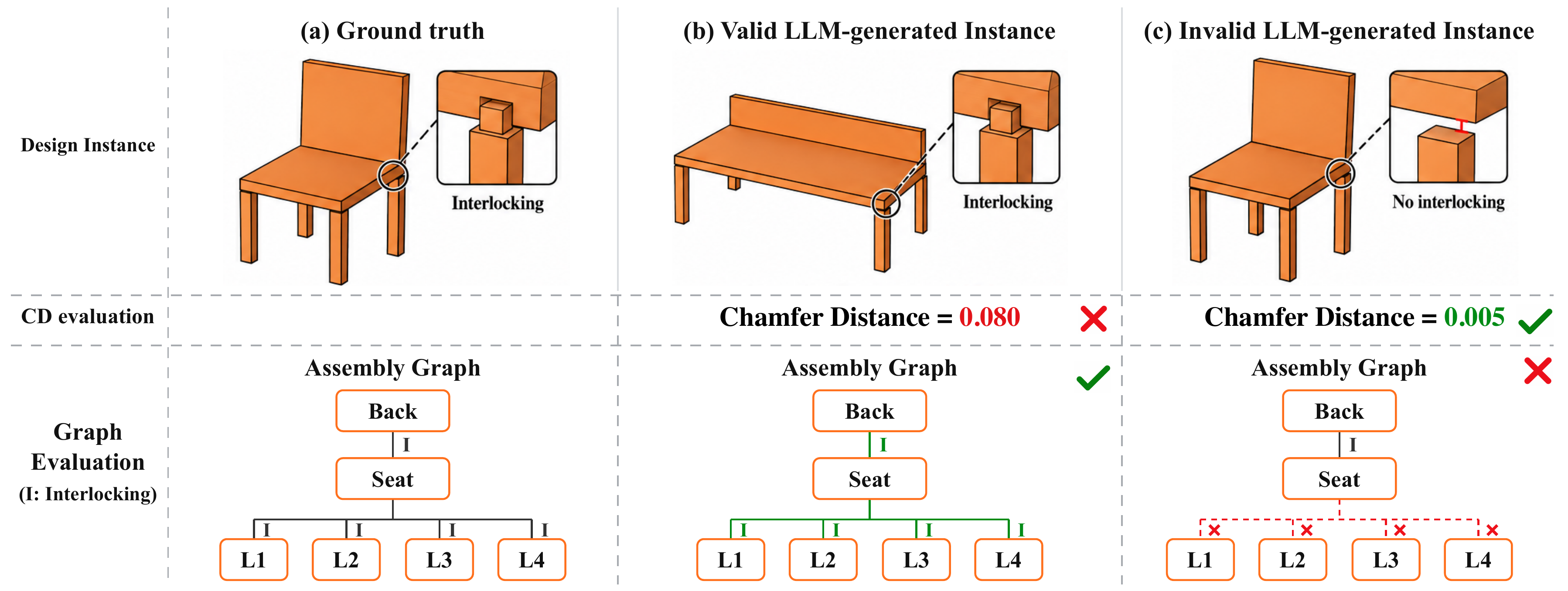}
  \caption{
    Chamfer Distance can misjudge multi-component CAD designs,
    while SIE evaluates assembly structure and interface geometry
    through programmatic STEP checks.
  }
  \label{fig:preliminary}
\end{figure}

Beyond dataset complexity, evaluation is another key challenge
for Text-to-CAD.
Existing evaluation methods can be broadly grouped into
three categories: \textit{geometric similarity},
\textit{visual alignment}, and
\textit{programmatic STEP checks}.
Geometric similarity metrics, such as Chamfer Distance
(CD)~\cite{Wang_2025_CADFusion}, measure shape resemblance
but may overlook errors in connectivity, interface types,
and interface clearances in multi-component designs.
For example, a functionally valid chair may receive a poor
CD score, while a geometrically similar one may fail to satisfy
these requirements (Figure~\ref{fig:preliminary}).
Visual alignment evaluates rendered CAD models against textual
or visual references using VLMs~\cite{articad}.
However, it requires additional model
inference and may produce unreliable judgments, particularly
when details such as interface types and interface clearances
are obscured in rendered views.
Programmatic STEP checks, as used in
CADTestBench~\citep{mallis2026cadtestbench}, directly inspect
generated STEP models but remain focused on single-part designs.

To address these limitations, we propose \model, a benchmark
for practical Text-to-CAD generation emphasizing
\textbf{assembly structure} and \textbf{interface geometry}
in complex multi-component designs.
Here, assembly structure denotes component configuration, while interface geometry denotes the geometric
features through which components connect and fit.
\model is constructed through collaboration between human
designers and LLMs.
Each design instance is defined by a structured
\textit{Design Specification} that describes the intended use,
manufacturing methods, admissible parameter ranges,
and requirements for assembly structure and interface geometry.
We further introduce an attributed graph representation,
the Assembly Graph, that explicitly records components,
connectivity, interface types, and interface clearances.
Each design instance is also paired with program-generated
CAD drawings for visual review.
All design instances are reviewed by human designers.

Our evaluation framework assesses generated CAD models along
five complementary dimensions: code executability, geometric
validity, structural and interface fidelity, and functional
consistency.
To evaluate assembly structure and interface geometry, we
introduce \textbf{Structure and Interface Evaluation (SIE)},
which uses programmatic STEP checks guided by the reference
Assembly Graph.
SIE evaluates four aspects: Structure Correspondence,
connectivity, interface types, and interface clearances.
It first matches generated components to the design reference,
then evaluates the remaining three aspects on the matched
components.
By grounding these assessments in direct B-Rep measurements,
SIE provides explicit geometric evidence for the fidelity
of assembly structure and interface geometry.
For these dimensions, its deterministic programmatic STEP
checks avoid the inference latency and token costs associated
with VLM-based visual alignment.
Functional consistency, which is less amenable to direct
geometric verification, is evaluated separately using a
VLM-as-judge protocol.

Experiments on \model reveal that both proprietary and
open-weight LLMs remain far from reliable end-to-end CAD
generation.
Even the strongest evaluated model, GPT-6 Astra, achieves only
57.55\% joint success, with substantial residual errors in
geometric validity, component recovery, assembly connectivity,
and interface typing.
Our analysis further shows that successful code execution is
not equivalent to correct CAD modeling, and that correctly
identified interface types do not necessarily imply accurate
interface clearances.
Together, these results identify geometric validity, assembly
structure, interface correctness, and clearance accuracy as key
remaining bottlenecks for robust multi-component Text-to-CAD
generation.

%% file: chapter/2.Related_Work.tex
\section{Related Work}

\textbf{Text-to-CAD Datasets.}
Text-to-CAD translates natural language into editable CAD models
through command-sequence prediction~\citep{Khan_2024_Text2CAD,Li_2024_CADTranslator}
or executable code generation~\citep{articad}.
Existing datasets commonly augment CAD construction sequences
with textual descriptions. For example, Text2CAD enriches
DeepCAD with automatically generated descriptions at different
levels of detail~\citep{Wu_2021_ICCV,Khan_2024_Text2CAD},
while CADFusion incorporates visual feedback to improve
text-to-shape alignment~\citep{Wang_2025_CADFusion}.
As summarized in Table~\ref{tab:text2cad_comparison},
these benchmarks largely emphasize single-part geometry,
with evaluation centered on geometric similarity and visual
alignment~\citep{Khan_2024_Text2CAD,Wang_2025_CADFusion}.

Although recent work such as ArtiCAD~\citep{articad} explores
assembly-aware generation, shape similarity alone does not
establish whether components are correctly separated,
connected, or dimensioned at mating interfaces.
A visually similar model may fuse distinct components or
introduce incorrect interface clearances.
In contrast, \model pairs multi-component B-Rep designs with
structured Design Specifications and evaluates component
correspondence, connectivity, interface types, and interface
clearances through direct geometric analysis.
These assessments are complemented by execution,
geometric-validity, and VLM-based functional checks.

\paragraph{Text-to-CAD Evaluation.}
Existing Text-to-CAD evaluation mainly relies on geometric
similarity, visual alignment, and programmatic STEP checks.
Geometric similarity metrics such as Chamfer Distance compare
generated and reference shapes~\cite{Wu_2021_ICCV,Khan_2024_Text2CAD},
while recent work also considers topology-aware
measures~\cite{preintner2025evocad}.
However, these metrics do not explicitly evaluate assembly
structure and interface geometry.
Visual alignment using VLMs has also been adopted for CAD
verification and refinement~\cite{Khan_2024_Text2CAD,alrashedy2024generating,cadsmith_2026},
but is limited by occlusion and viewpoint coverage, especially
for internal interface geometry and small interface clearances. More recently, executable tests directly on CAD geometry have emerged as a deterministic alternative~\cite{mallis2026cadtests}. Our evaluation also operates directly on CAD geometry, but targets a different aspect: component correspondence, typed interface connectivity, and local interface clearances in multi-component CAD designs.

\input{table/preliminary}

%% file: table/preliminary.tex
\begin{table}[t]
\centering
\caption{Comparison of Representative Existing Text-to-CAD Benchmarks.}
\label{tab:text2cad_comparison}
\resizebox{\linewidth}{!}{
\begin{tabular}{cccc}
\toprule
\textbf{Benchmark} &
\textbf{Input Type} &
\textbf{Dataset Complexity} &
\textbf{Evaluation Focus} \\
\midrule

Text2CAD~\citep{khan2024text2cad}
&
Modeling sequence
&
Single-part
&
Geometric similarity 
\\

CADPrompt~\citep{alrashedy2024generating}
&
Text description
&
Single-part
&
Geometric similarity 
\\

Text2CAD-Bench~\citep{wang2026text2cadbench}
&
Geo. / Seq. description
&
Single-part
&
Geometric similarity
\\

CADTestBench~\citep{mallis2026cadtestbench}
&
Text requirements
&
Single-part
&
Programmatic STEP Checks
\\

ArtiCAD~\citep{articad}
&
Text / image description
&
Multi-part assembly
&
Visual Alignment
\\

\textbf{\model (Ours)}
&
\textbf{Design description \& assembly graph}
&
\textbf{Multi-part assembly}
&
\textbf{Programmatic STEP Checks }
\\

\bottomrule
\end{tabular}
}
\end{table}

%% file: chapter/3.Benchmark.tex
\begin{figure}[htbp]
  \centering
  \includegraphics[width=1\linewidth]{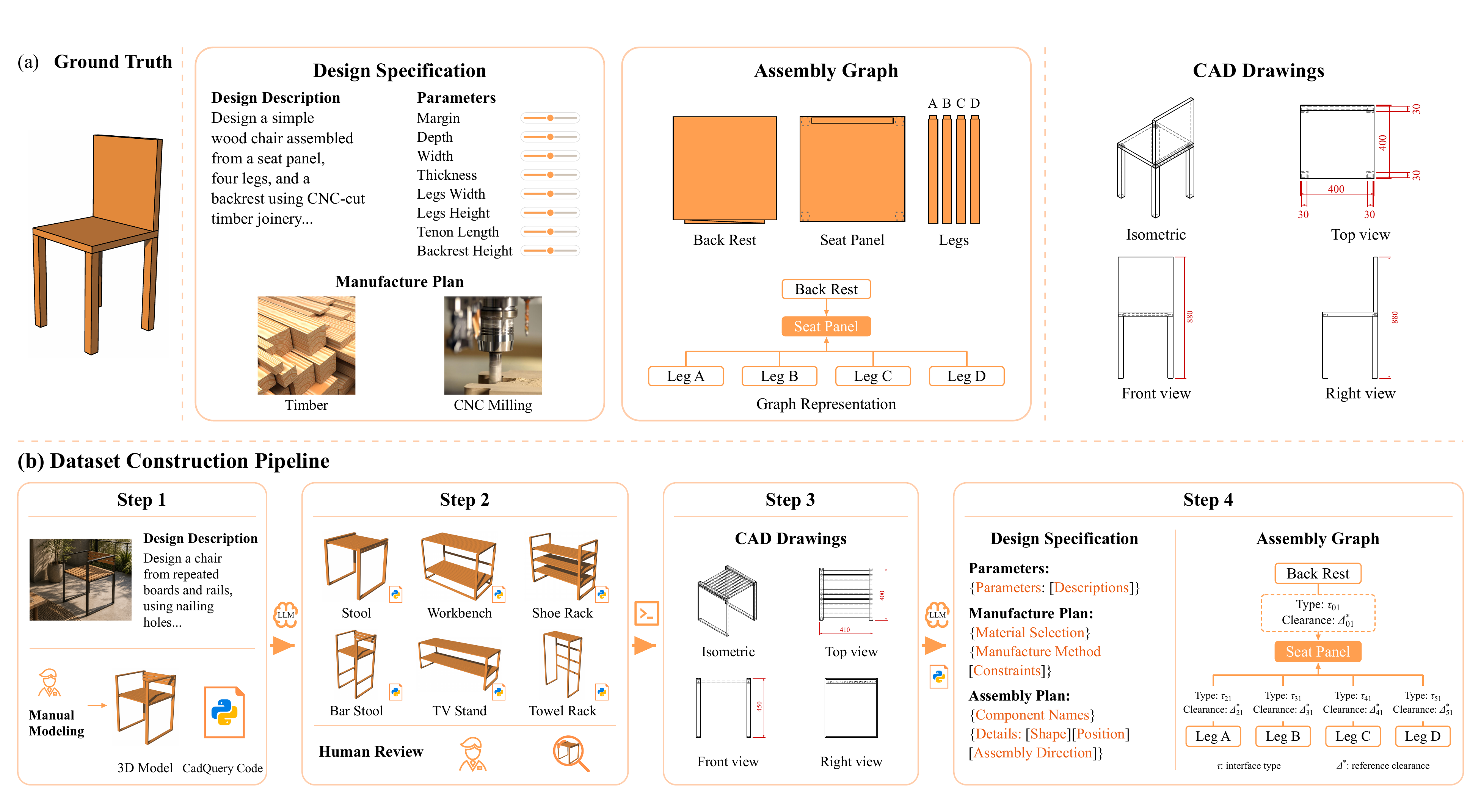}
  \caption{
  Overview of our dataset construction.
  (a) Each benchmark instance provides a Design Specification, assembly graph and CAD drawings.
  (b) The dataset is curated through expert seed modeling,
  LLM-based augmentation, CAD drawing generation, and
  Design Specification synthesis.
  }
  \label{fig:dataset_construction}
\end{figure}
\section{\model: Benchmark Design}
\subsection{Preliminaries: Design Description and Assembly Graph}

Existing Text-to-CAD benchmarks primarily use text descriptions
or images as inputs, as summarized in Table~\ref{tab:text2cad_comparison}, which do not explicitly represent assembly
structure and interface geometry.
We therefore introduce a Physical Assembly Graph to make this
information explicit for multi-component CAD designs. Each benchmark instance provides two inputs, organized as a
structured \textit{Design Specification}:
\begin{equation}
\mathcal S=\langle\mathcal D,\mathcal G\rangle,
\end{equation}
where $\mathcal D$ is a natural-language design description and
$\mathcal G$ is a Physical Assembly Graph.
The description $\mathcal D$ specifies the intended use,
manufacturing methods, admissible parameter ranges.
The graph $\mathcal G$ explicitly represents assembly structure and interface geometry.

\begin{definition}[Physical Assembly Graph]
A Physical Assembly Graph is an attributed undirected graph:
\begin{equation}
\mathcal G=(\mathcal V,\mathcal E),
\qquad
\mathbf a(v_i)=\ell_i,
\qquad
\mathbf a(e_{ij})=
\langle\tau_{ij},\Delta^{*}_{ij}\rangle,
\end{equation}
where each node $v_i\in\mathcal V$ represents a physical
component with a semantic label $\ell_i$, such as
\emph{seat panel} or \emph{support leg}.
Each edge $e_{ij}\in\mathcal E$ represents a physical interface
between components $v_i$ and $v_j$, with interface type
$\tau_{ij}$ and, where applicable, a reference interface clearance
$\Delta^{*}_{ij}$.
Each reference interface clearance $\Delta^{*}_{ij}$ is
accompanied by an allowable range $[L_{ij},U_{ij}]$,
which specifies the lower and upper bounds for the
generated interface clearance.
\end{definition}

\subsection{Dataset Construction Pipeline}
\label{sec:dataset_construction}
We introduce a dataset of strictly functional, manufacturable and assemblable CAD models. We employ a ``Human-in-the-loop'' pipeline that synergizes expert domain knowledge with LLM-driven scalable augmentation.
The dataset curation pipeline consists of four stages, as illustrated in Figure~\ref{fig:dataset_construction}. We first create high-quality seed 3D models with expert designers, then expand them into diverse 3D model variants using LLM-assisted CAD script augmentation. The resulting models are converted into standardized engineering views, and finally paired with structured Design Specifications.

\noindent\textbf{Step 1: Expert-Guided Construction of Seed Models.}
We first collect the design description and rendered reference examples of established designs across diverse design categories (e.g., chairs, tables, and business card holders). Following these instructions, professional designers manually construct high-quality 3D models in STEP format as the seed models. We then convert the designers' modeling procedures into executable \textit{CadQuery} scripts, which serve as the basis for subsequent augmentation.

\noindent\textbf{Step 2: LLM-Powered Data Augmentation.} 
To maximize dataset diversity, we use Claude Opus 4.7 to systematically expand the initial \textit{CadQuery} scripts along two primary dimensions: \textit{stylistic transformations} (e.g., adapting a traditional wooden chair into modern industrial aesthetics) and \textit{functional repurposing} (e.g., transforming a chair's into a shoe rack or coat stand).

\noindent\textbf{Step 3: Engineering View and Physical Assembly Graph Generation.}
For each seed and augmented model, we generate engineering
views and a Physical Assembly Graph $\mathcal G$.
To mitigate occlusion and limited viewpoint coverage in
conventional rendered images~\cite{pointercad}, we use the
geometry kernel to extract boundaries, contours, and hidden
lines, with hidden lines shown as dashed paths.
The views are arranged in a standardized $2\times2$ grid
(top, front, right, and isometric) for visual inspection
(Figure~\ref{fig:dataset_construction}(a)).
We also provide each STEP model to an LLM to generate
$\mathcal G$, recording components, connectivity, interface
types, and reference interface clearances with their
allowable ranges.

\noindent\textbf{Step 4: Design Specification Synthesis and Expert Review.}
Using the engineering views, \textit{CadQuery} scripts,
and Physical Assembly Graphs, we prompt GPT-5.5 with
expert-written examples (Appendix~\ref{app:prompt_templates})
to generate design descriptions $\mathcal D$ covering
intended use, manufacturing methods, admissible parameter
ranges, and interface clearance requirements.
Each description is paired with $\mathcal G$ to form
$\mathcal S=\langle\mathcal D,\mathcal G\rangle$.
Human experts review and correct both $\mathcal D$ and
$\mathcal G$ for consistency with the CAD model and
design requirements.


\subsection{Evaluation Protocol}
\label{sec:evaluation}

\begin{figure}[htbp]
  \centering  \includegraphics[width=1.0\linewidth]{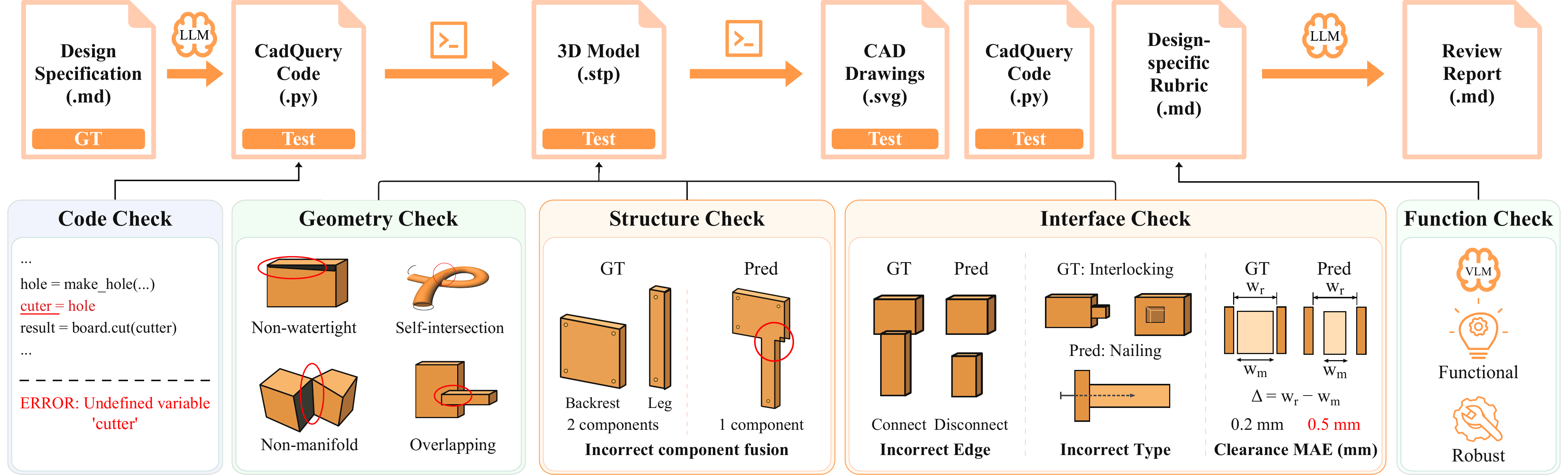}
  \caption{
  Illustration of the proposed evaluation pipeline. Given a Design Specification, each evaluated model generates a \textit{CadQuery} script, which is executed and exported as a STEP file. The evaluation then performs code, geometry, structure, interface, and function checks.
  }
  \label{fig:evaluation}
\end{figure}

Given a Design Specification
$\mathcal S=\langle\mathcal D,\mathcal G\rangle$,
each model generates a \textit{CadQuery} program, which is
executed to produce a STEP model $C$ and standardized
engineering drawings.
We evaluate the outputs through five checks:
code, geometry, structure, interface, and function,
as illustrated in Figure~\ref{fig:evaluation} and summarized
in Table~\ref{tab:overall-metrics-gemini}.
The code check assesses program executability;
the geometry, structure, and interface checks operate
on the STEP model; and the function check uses a VLM
to assess the engineering drawings.

\textbf{Code Check.}
A sample passes \textbf{Sandbox Success} if its
\textit{CadQuery} program terminates without execution errors,
constructs a non-empty CAD model, and successfully exports
a STEP file.
This check measures whether the model produces an executable
CAD artifact.

\textbf{Geometry Check.}
For exported STEP models, we evaluate four independent binary
criteria.
\textit{Watertight} requires closed solids without open
boundaries; \textit{Manifold} requires valid manifold topology;
\textit{Self-Intersection Free} excludes self-intersecting faces
or invalid internal intersections within individual solids;
and \textit{Overlap Free} excludes physical interpenetration
between distinct components.
\textit{Geom.\ Valid} requires successful execution and passes
on all four checks.

\textbf{Structure Check.}
Geometric validity alone does not establish whether a model
contains the intended physical components.
Let the reference Physical Assembly Graph be
$\mathcal G^*=(\mathcal V^*,\mathcal E^*)$.
SIE associates generated solids with labeled reference
components using normalized relative-position, dimension,
and face-count descriptors.
An augmented Hungarian assignment permits unmatched components;
mutual-nearest and ambiguity-margin checks reject uncertain
correspondences.
The full procedure appears in Algorithm~\ref{alg:naming}
in Appendix~\ref{app:evaluation_details}.
We compare the resulting labeled node set
$\widehat{\mathcal V}$ with $\mathcal V^*$ using
\textit{Node Precision}, \textit{Node Recall}, and
\textit{Node F1}.
A correct node requires both component presence and correct
semantic correspondence.

\noindent\textbf{Interface Check.}
SIE further evaluates connectivity, interface types,
and interface clearances through the following three checks.

\noindent\textit{(1) Connectivity and interface type checks.}
We detect connections between solids and identify
their interface types, e.g., interlocking (Algorithm~2 in Appendix~\ref{app:evaluation_details}).
As an example of interlocking detection, rectangular
mortise-and-tenon interfaces are identified from nearby,
oppositely oriented planar faces with overlapping projections.
Four-wall candidates are verified using orthogonality,
common-section, and solid-containment checks, and retained
when all four male walls match inward-facing receiver walls.
For connectivity, the detected component pairs define the
predicted edge set $\widehat{\mathcal E}$, which is compared
with the reference edge set $\mathcal E^\ast$ using
\textit{Edge F1}.
For interface type checks, we evaluate whether the detected
connections have the correct interface types using
\textit{Type F1}.


\noindent\textit{(2) interface clearances.}
For each localized rectangular interface $e$, we recover
paired male and receiver side faces from the STEP geometry.
Their single-side clearance is
\begin{equation}
g_{ei}
=
(\mathbf c^r_{ei}-\mathbf c^m_{ei})^\top
\mathbf n^m_{ei},
\end{equation}
where $\mathbf c^m_{ei}$ and $\mathbf c^r_{ei}$ are face
centroids and $\mathbf n^m_{ei}$ is the male face's outward
normal.
The four side clearances form two opposing pairs, giving
total clearances along two cross-sectional directions:
\begin{equation}
\Delta_{e1}=g_{e1}+g_{e2},
\qquad
\Delta_{e2}=g_{e3}+g_{e4}.
\end{equation}
Equivalently, $\Delta_{ek}=w^r_{ek}-w^m_{ek}$, where
$k\in\{1,2\}$ indexes the cross-sectional directions, and
$w^r_{ek}$ and $w^m_{ek}$ are receiver and male widths.
Agreement between these calculations provides an internal
consistency check.
We measure deviation from reference total clearances
$\Delta^*_{ek}$ using
\begin{equation}
\operatorname{ClearanceMAE}
=
\frac{1}{N_\Delta}
\sum_{(e,k)\in\mathcal Q}
\left|\Delta_{ek}-\Delta^*_{ek}\right|,
\end{equation}
where $\mathcal Q$ contains interface--direction pairs
with both predicted and reference measurements, and
$N_\Delta=|\mathcal Q|$.
Clearance MAE is reported in millimeters and measures
reference-clearance deviation, not tolerance compliance.%
\footnote{Clearance MAE evaluates only matched, measurable
interface clearances, excluding unavailable measurements
rather than assigning them zero.}
Compliance with the specified allowable range
$[L_{ek},U_{ek}]$ requires
$L_{ek}\leq\Delta_{ek}\leq U_{ek}$.


\textbf{Function Check.}
Even a geometrically valid model with correct component
structure may fail to satisfy its intended function.
A VLM therefore evaluates the generated SVG engineering
views together with the Design Specification $\mathcal S$.
\textbf{Functional} assesses whether the design contains
the structures needed for its intended use.
\textbf{Robust} assesses whether the visible component
arrangement, supports, interfaces, and apparent load-transfer
structure suggest a mechanically plausible and stable design.
These criteria are scored independently and represent
visual assessments rather than physical validation.
The evaluation prompt and scoring protocol are provided
in Appendix~\ref{app:prompt_templates}.

%% file: chapter/4.Experiments.tex
\section{Experiments}
We first provide an overview of \model. We then use our five-stage evaluation pipeline to assess the CAD models generated by proprietary and open-source LLMs.



\subsection{Data Distribution of \model}

\begin{figure}[htbp]
  \centering\includegraphics[width=1\linewidth]{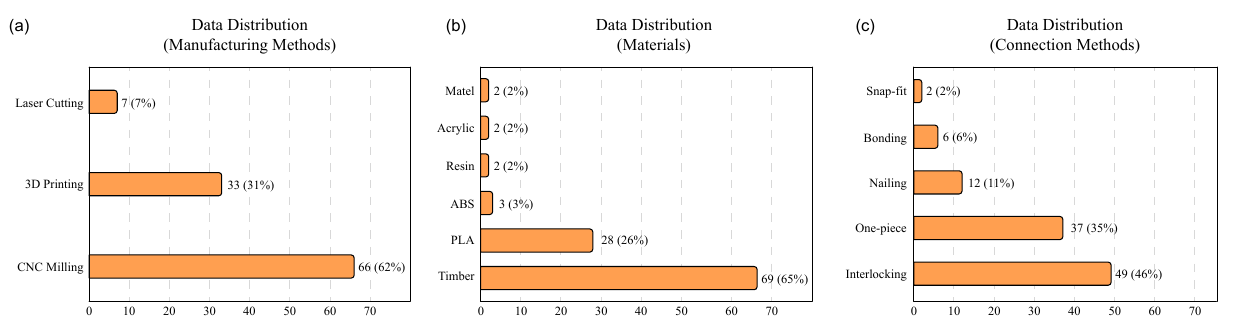}
\caption{
Dataset statistics of \model, showing the distributions of (a) manufacturing methods, (b) materials, and (c) connection methods across all design instances.
}
  \label{fig:data_statistics}
\end{figure}


\model comprises 106 design instances spanning a wide range of manufacturing processes, materials, and connection methods, as summarized in Figure~\ref{fig:data_statistics}. The benchmark includes representative manufacturing processes such as computer numerical control (CNC) milling, 3D printing, and laser cutting, with CNC milling constituting the largest category. In terms of material usage, timber and polylactic acid filaments (PLA) are the most prevalent. Regarding connection methods, one-piece designs represent single-body objects, whereas interlocking, nailing, and snap-fit designs correspond to multi-component assemblies. Overall, these distributions indicate that \model places particular emphasis on assemblable CAD models while maintaining broad coverage of real-world manufacturable design scenarios.

\input{table/main_experiments_gemini}

\subsection{Evaluation of Closed-Source and Open-Source LLMs}
\label{sec:experiments}
We evaluate a broad set of closed-source and open-source LLMs
using the proposed five-stage evaluation pipeline.
Given a Design Specification $\mathcal S$, each model generates
a \textit{CadQuery} script, evaluated through
(1) code, (2) geometry, (3) structure, (4) interface,
and (5) function checks.
The code check assesses program executability;
the geometry, structure, and interface checks assess the
resulting STEP model; and the function check evaluates
standardized engineering drawings using a VLM.
The structure and interface checks jointly constitute SIE,
assessing assembly structure and interface geometry.
The closed-source models evaluated include GPT-6 Astra%
\footnote{GPT-6 Astra is accessed programmatically through
Codex, whereas all other models are accessed through
OpenRouter. All models receive the same task inputs
under a single-pass generation setting.},
GPT-5.5, GPT-4o, Claude Opus 4.7, Claude 3.7 Sonnet,
and Gemini 3.1 Pro.
The open-source models include GLM-5.1, GLM-4.7-Flash,
MiniMax-M2.7, MiniMax-M2.5, Qwen3.5-122B-A10B,
Qwen2.5-72B, Llama-3.1-70B, Qwen3.6-35B-A3B,
Qwen3.6-Coder, and Llama-3.1-8B.

\textbf{Inapplicability of Existing Text-to-CAD Baselines.} We do not directly evaluate existing Text-to-CAD methods, as their task settings differ fundamentally from ours and are not applicable to our benchmark. Prior work primarily uses command-sequence-style inputs to generate isolated primitive or single-part CAD models~\citep{xu2024cad,liao2025automated,Yavartanoo_2024_Text2CAD,alrashedy2024generating,pointercad}. These methods are architecturally and procedurally tailored to such data and interaction paradigms, and thus cannot be directly applied to our design instances, which are manufacturable multi-component assemblies with explicit component relations and manufacturing constraints. ArtiCAD~\citep{articad} addresses assembly-level CAD generation,
but we exclude it from our experiments because its implementation
was unavailable at the time of evaluation.



\subsection{Experimental Results and Analysis}
\label{sec:results_analysis}

Table~\ref{tab:overall-metrics-gemini} summarizes performance
across the five checks defined in our evaluation protocol.
For cases that fail code execution, the per-case precision,
recall, and F1 scores for components, connectivity, and
interface types are set to zero. These scores are averaged
over all 106 cases. We additionally report \textbf{Joint Success}:
a case receives a score of one only if it passes all five
checks under the criteria defined in the evaluation protocol,
and zero otherwise.
Table~\ref{tab:assembly-graph-diagnostics} provides a more
detailed breakdown of assembly-graph evaluation, reporting
precision and recall for components, connectivity, and
interface types. The STEP column counts cases with available
STEP outputs.
We also report \textit{Graph Edit Distance (GED)}, the minimum
node and edge edit cost required to transform the predicted
assembly graph into the reference graph.

\paragraph{RQ1: What are the main bottlenecks?}
Geometry, assembly structure, and interface correctness remain
major bottlenecks, with substantial performance gaps relative
to code execution success
(Table~\ref{tab:overall-metrics-gemini}).
For example, Claude Opus 4.7 achieves 76.42\% execution success,
but only 58.49\% geometric validity, 63.77\% Node F1,
and 33.02\% Type F1.
The gaps are even larger for Qwen-2.5-72B:
despite 50.00\% execution success, it reaches only
19.81\% geometric validity, 20.58\% Node F1,
and 0.00\% Type F1.
\textbf{Even GPT-6 Astra still falls short.}
Although it leads these metrics, its 99.06\% execution success
is accompanied by 89.62\% geometric validity,
89.77\% Node F1, 86.08\% Edge F1, and 74.22\% Type F1,
leaving room for improvement in component recovery,
connectivity, and interface typing.

\paragraph{RQ2: Does strong code generation guarantee correct CAD models?}
High execution success does not guarantee end-to-end CAD correctness.
GPT-6 Astra executes successfully on 99.06\% of designs,
yet achieves only 57.55\% joint success,
meaning that only slightly more than half of all designs
satisfy the combined evaluation requirements.
The same gap appears in other models:
GPT-5.5 achieves 77.36\% execution success but
34.91\% joint success, while Claude Opus 4.7 achieves
76.42\% and 23.58\%, respectively.
Thus, producing executable code remains distinct from
generating models that jointly satisfy geometric,
structural, interface, and functional requirements.

\paragraph{RQ3: Are correctly typed interfaces properly fitted?}
Clearance MAE captures dimensional interface clearance errors that
interface type scores alone cannot reveal, and the two
metrics do not consistently rank models in the same order.
For example, GPT-5.5 achieves the second-highest Type F1
(48.64\%) but the largest reported clearance MAE
(0.4750\,mm).
This illustrates that correctly typed interfaces may still fit poorly for practical use.
Conversely, MiniMax-M2.7 has the lowest clearance MAE
(0.2000\,mm). Since this metric is computed only on detected,
measurable connections, its low error may be attributed to the simpler
interface types it generates, which are easier to fit.

\begin{figure*}[t]
    \centering
    \includegraphics[width=\textwidth]{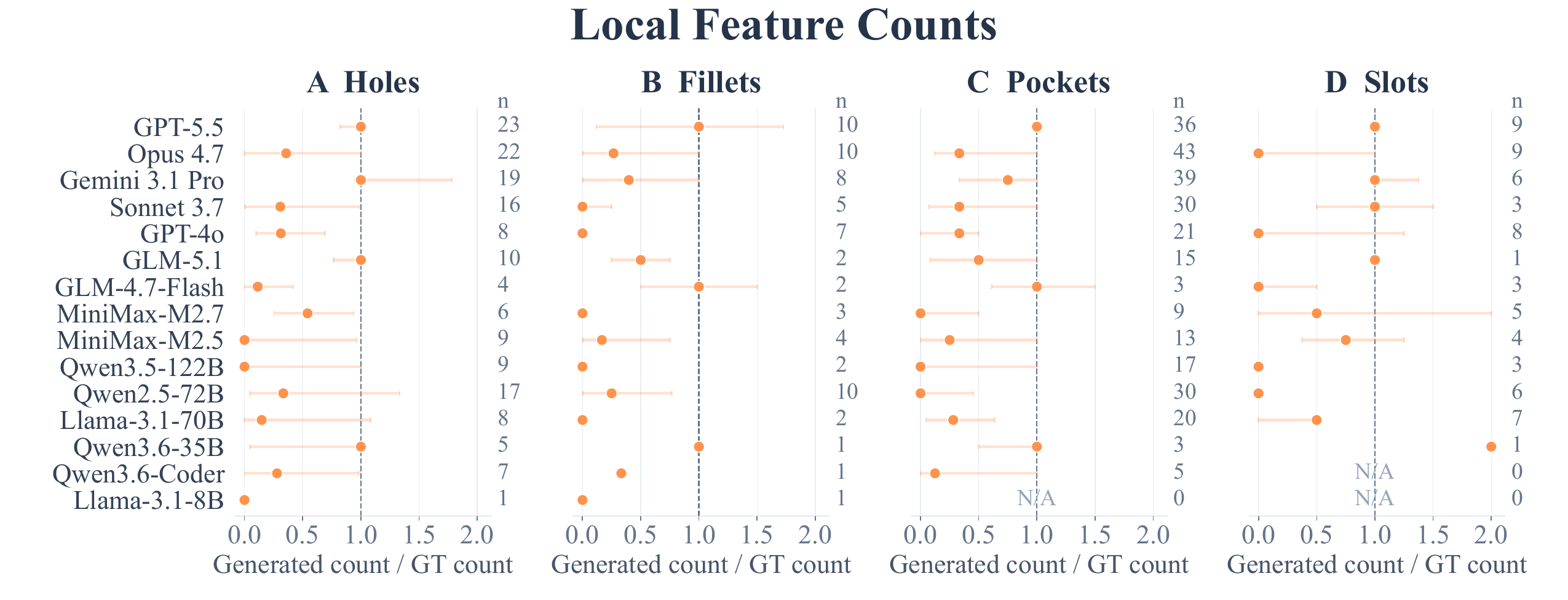}
    \caption{\textbf{Local feature counts.}
    Dots and light bars show the median and 25th--75th percentiles
    of per-design generated-to-GT count ratios.
    $n$ denotes eligible designs.
    Details are in Appendix~\ref{app:local-geometric-analysis}.}
    \label{fig:local-feature-counts}
\end{figure*}

\paragraph{Local Geometric Feature Analysis.}
We analyze local geometric features to characterize how
LLM-generated CAD models deviate from reference designs,
considering both feature counts (holes, fillets, pockets, and slots)
and feature dimensions (diameters, depths, and widths).
Definitions and computation details are provided in
Appendix~\ref{app:local-geometric-analysis}.
For each eligible design, we normalize each generated descriptor
by its ground-truth counterpart.
Figures~\ref{fig:local-feature-counts}
and~\ref{fig:local-feature-dimensions} summarize these ratios,
with dots indicating medians and light bars showing the
25th--75th percentile ranges.
A ratio of 1 indicates numerical agreement with the reference,
while ratios below or above 1 indicate smaller or larger values,
respectively.
Most models have median hole-count ratios below 1,
indicating a tendency to generate fewer holes than the references.
Agreement in feature counts, however, does not necessarily
extend to feature dimensions, and dimensional agreement
can vary across descriptors.
For GPT-5.5, Opus, and Gemini, hole-diameter ratios
concentrate near 1, whereas the interquartile ranges of
hole-depth ratios extend toward lower values.
GPT-5.5 illustrates this distinction particularly clearly:
its median count ratio is 1 across all four feature categories,
yet it produces shallower holes in some designs despite
close agreement in hole diameters.

\input{table/iaa}

\subsection{Human Alignment of the VLM Judge}
\label{sec:llm-human-agreement}

We assess VLM--human agreement for the function check using
rendered outputs from 20 sampled design instances, annotated
by four raters across two rounds. Human scores are obtained
by averaging available annotations for the Functional and
Robust sub-criteria. We compare these scores with VLM
judgments using Pearson, Spearman, and Kendall correlations.

%% file: table/main_experiments_gemini.tex

\begin{table*}[t]
\centering
\begingroup
\caption{Performance across five evaluation checks. Bold and italics indicate the best and second-best available values, respectively, including ties.}
\label{tab:overall-metrics-gemini}

\renewcommand{\arraystretch}{1.15}
\setlength{\tabcolsep}{3pt}

\resizebox{\textwidth}{!}{%
\begin{tabular}{l*{13}{c}}
\toprule
& \multicolumn{1}{c}{\textbf{Code Check}}
& \multicolumn{5}{c}{\textbf{Geometry Check}}
& \multicolumn{1}{c}{
    \textbf{Structure Check}}
& \multicolumn{3}{c}{
    \textbf{Interface Check}}
& \multicolumn{2}{c}{\textbf{Function Check}} & \multicolumn{1}{c}{\textbf{Overall}} \\
\cmidrule(lr){2-2}
\cmidrule(lr){3-7}
\cmidrule(lr){8-8}
\cmidrule(lr){9-11}
\cmidrule(lr){12-13}\cmidrule(lr){14-14}

Model
& \shortstack{Sandbox\\Success $\uparrow$}
& Watertight $\uparrow$
& Manifold $\uparrow$
& \shortstack{Self-Int.\\Free $\uparrow$}
& \shortstack{Overlap\\Free $\uparrow$}
& \shortstack{Geom.\\Valid $\uparrow$}
& Node F1 $\uparrow$
& Edge F1 $\uparrow$
& Type F1 $\uparrow$
& \shortstack{Clearance\\MAE$\downarrow$}
& Functional $\uparrow$
& Robust $\uparrow$ & \shortstack{Joint\\Success $\uparrow$} \\
\midrule

\rowcolor{orange!10}
\multicolumn{14}{c}{\textit{Closed-Source Models}} \\
\rowcolor{gray!8}
gpt-6-astra & \textbf{99.06} & \textbf{97.17} & \textbf{97.17} & \textbf{89.62} & \textbf{97.17} & \textbf{89.62} & \textbf{89.77} & \textbf{86.08} & \textbf{74.22} & 0.3341 & \textbf{86.79} & \textbf{84.91} & \textbf{57.55} \\
claude-opus-4.7 & 76.42 & \textit{76.42} & \textit{76.42} & \textit{73.58} & 60.38 & 58.49 & 63.77 & 48.39 & 33.02 & 0.3082 & 43.40 & 42.45 & 23.58 \\
\rowcolor{gray!8}
claude-3.7-sonnet & 46.23 & 46.23 & 46.23 & 46.23 & 23.58 & 23.58 & 30.18 & 11.96 & 8.20 & \textit{0.2353} & 11.32 & 13.21 & 8.49 \\
gemini-3.1-pro & 65.09 & 59.43 & 63.21 & 63.21 & 62.26 & 56.60 & 53.52 & 53.66 & 41.24 & 0.3066 & 47.17 & 45.28 & 22.64 \\
\rowcolor{gray!8}
gpt-5.5 & \textit{77.36} & 72.64 & 74.53 & \textit{73.58} & \textit{70.75} & \textit{66.04} & \textit{69.05} & \textit{59.19} & \textit{48.64} & 0.4750 & \textit{51.89} & \textit{52.83} & \textit{34.91} \\
gpt-4o & 36.79 & 36.79 & 36.79 & 36.79 & 13.21 & 13.21 & 17.78 & 4.58 & 1.41 & -- & 1.89 & 5.66 & 1.89 \\
\midrule
\rowcolor{orange!10}
\multicolumn{14}{c}{\textit{Open-Source Models}} \\
\rowcolor{gray!8}
glm-5.1 & 31.13 & 31.13 & 31.13 & 31.13 & 27.36 & 27.36 & 29.12 & 17.93 & 14.64 & 0.3641 & 17.92 & 19.81 & 15.09 \\
glm-4.7-flash & 8.49 & 7.55 & 7.55 & 7.55 & 5.66 & 4.72 & 5.25 & 0.00 & 0.00 & -- & 1.89 & 1.89 & 0.94 \\
\rowcolor{gray!8}
minimax-m2.7 & 18.87 & 18.87 & 18.87 & 18.87 & 10.38 & 10.38 & 13.26 & 2.22 & 2.22 & \textbf{0.2000} & 5.66 & 4.72 & 4.72 \\
minimax-m2.5 & 25.47 & 22.64 & 22.64 & 22.64 & 10.38 & 9.43 & 11.23 & 1.84 & 0.21 & -- & 1.89 & 1.89 & 1.89 \\
\rowcolor{gray!8}
qwen-3.5-122b-a10b & 27.36 & 25.47 & 25.47 & 25.47 & 13.21 & 11.32 & 15.56 & 2.42 & 2.42 & -- & 6.60 & 8.49 & 4.72 \\
qwen-2.5-72b & 50.00 & 49.06 & 49.06 & 49.06 & 19.81 & 19.81 & 20.58 & 2.61 & 0.00 & -- & 1.89 & 3.77 & 0.94 \\
\rowcolor{gray!8}
llama-3.1-70b & 31.13 & 30.19 & 30.19 & 30.19 & 22.64 & 22.64 & 8.78 & 0.00 & 0.00 & -- & 0.00 & 1.89 & 0.00 \\
qwen-3.6-35b-a3b & 8.49 & 8.49 & 8.49 & 8.49 & 5.66 & 5.66 & 6.57 & 0.18 & 0.00 & -- & 2.83 & 4.72 & 2.83 \\
\rowcolor{gray!8}
qwen-3.6-coder-next & 14.15 & 14.15 & 14.15 & 14.15 & 10.38 & 10.38 & 9.37 & 0.00 & 0.00 & -- & 2.83 & 4.72 & 1.89 \\
llama-3.1-8b & 2.83 & 1.89 & 1.89 & 1.89 & 0.94 & 0.94 & 0.63 & 0.00 & 0.00 & -- & 0.00 & 0.00 & 0.00 \\
\bottomrule
\end{tabular}%
}

\endgroup
\end{table*}

%% file: table/iaa.tex
\begin{table*}[t]
\centering
\caption{VLM--human agreement on Functional and Robust scores.}
\label{tab:judge-corr}
\setlength{\tabcolsep}{4pt}
\renewcommand{\arraystretch}{1.15}
\resizebox{\textwidth}{!}{%
\begin{tabular}{l ccc ccc ccc}
\toprule
 & \multicolumn{3}{c}{Gemini 3.1 Pro} & \multicolumn{3}{c}{GPT-5.5} & \multicolumn{3}{c}{GPT-4o} \\
\cmidrule(lr){2-4}\cmidrule(lr){5-7}\cmidrule(lr){8-10}
Level & Pearson $r$ & Spearman $\rho$ & Kendall $\tau$
& Pearson $r$ & Spearman $\rho$ & Kendall $\tau$
& Pearson $r$ & Spearman $\rho$ & Kendall $\tau$ \\
\midrule
Sub-criteria
& 0.718 & 0.699 & 0.648
& 0.653 & 0.661 & 0.601
& 0.626 & 0.603 & 0.572 \\
Criteria
& 0.755 & 0.756 & 0.674
& 0.719 & 0.698 & 0.632
& 0.696 & 0.692 & 0.604 \\
Design instance
& 0.842 & 0.824 & 0.701
& 0.768 & 0.781 & 0.647
& 0.819 & 0.807 & 0.670 \\
\bottomrule
\end{tabular}}
\end{table*}

%% file: chapter/5.Conclusion.tex
\section{Conclusion}
\label{sec:conclusion}



We introduced \model, a benchmark for multi-component
Text-to-CAD generation with explicit structural and fit
requirements. Its Structure and Interface Evaluation (SIE) combines reference
assembly graphs with direct B-Rep analysis to assess Structure
Correspondence, connectivity, interface types, and interface
clearances, while functional consistency is evaluated separately
using a VLM. Our results show that generating
executable CAD models does not guarantee correct assembly
structure or interface geometry, highlighting the need for
evaluation beyond global shape similarity.

\paragraph{Limitations.}
Our evaluation focuses on static assembly structure and
interface geometry, but does not evaluate the feasibility of
assembly sequences. Such evaluation requires reasoning about
component motion, collision-free insertion paths, and
intermediate assembly states, which are beyond the scope of
our current geometric checks. Moreover, the reference designs
have not been physically manufactured or assembled, and
VLM-based functional assessments cannot verify real-world
mechanical performance.

%% file: chapter/Appendix.tex
\input{chapter/evaluation_details}

\begin{table*}[t]
\centering
\begingroup
\caption{Component and connection graph evaluation. Bold and italics indicate the best and second-best available values, respectively, including ties.}
\label{tab:assembly-graph-diagnostics}

\renewcommand{\arraystretch}{1.15}
\setlength{\tabcolsep}{3pt}

\resizebox{\textwidth}{!}{%
\begin{tabular}{l*{12}{c}}
\toprule
& \multicolumn{1}{c}{\textbf{Coverage}}
& \multicolumn{3}{c}{\textbf{Components}}
& \multicolumn{3}{c}{\textbf{Connectivity}}
& \multicolumn{3}{c}{\textbf{Interface Types}}
& \multicolumn{2}{c}{\textbf{Component Graph}} \\
\cmidrule(lr){2-2}
\cmidrule(lr){3-5}
\cmidrule(lr){6-8}
\cmidrule(lr){9-11}
\cmidrule(lr){12-13}

Model
& STEP
& Cases & Precision $\uparrow$ & Recall $\uparrow$
& Cases & Precision $\uparrow$ & Recall $\uparrow$
& Cases & Precision $\uparrow$ & Recall $\uparrow$
& Cases & GED $\downarrow$ \\
\midrule

\rowcolor{orange!10}
\multicolumn{13}{c}{\textit{Closed-Source Models}} \\
\rowcolor{gray!8}
gpt-6-astra & 105 & 103 & \textbf{90.04} & \textbf{89.62} & 69 & \textbf{85.58} & \textbf{87.04} & 69 & \textbf{76.73} & \textbf{73.03} & 68 & \textit{0.4853} \\
claude-opus-4.7 & 81 & 81 & 63.82 & 64.78 & 59 & 57.66 & 45.02 & 59 & 39.71 & 30.27 & 44 & 2.0909 \\
\rowcolor{gray!8}
claude-3.7-sonnet & 49 & 49 & 29.50 & 32.09 & 33 & 15.36 & 11.01 & 33 & 10.51 & 7.56 & 25 & 1.9600 \\
gemini-3.1-pro & 69 & 69 & 53.10 & 54.56 & 50 & 53.38 & 54.40 & 50 & 42.93 & 40.44 & 37 & 1.4865 \\
\rowcolor{gray!8}
gpt-5.5 & 82 & 81 & \textit{68.90} & \textit{69.53} & 51 & \textit{60.48} & \textit{58.91} & 51 & \textit{51.71} & \textit{46.91} & 50 & 0.7200 \\
gpt-4o & 39 & 39 & 17.93 & 17.71 & 30 & 8.21 & 3.67 & 30 & 2.90 & 0.99 & 20 & 2.4500 \\
\midrule
\rowcolor{orange!10}
\multicolumn{13}{c}{\textit{Open-Source Models}} \\
\rowcolor{gray!8}
glm-5.1 & 33 & 33 & 28.96 & 29.44 & 17 & 18.77 & 17.97 & 17 & 15.46 & 14.12 & 24 & 1.2500 \\
glm-4.7-flash & 9 & 9 & 5.19 & 5.97 & 3 & 0.00 & 0.00 & 3 & 0.00 & 0.00 & 8 & 1.7500 \\
\rowcolor{gray!8}
minimax-m2.7 & 20 & 20 & 12.83 & 13.91 & 13 & 2.90 & 1.98 & 13 & 2.90 & 1.98 & 15 & 2.6000 \\
minimax-m2.5 & 27 & 27 & 10.88 & 11.94 & 20 & 3.02 & 1.34 & 20 & 0.48 & 0.13 & 10 & 2.4000 \\
\rowcolor{gray!8}
qwen-3.5-122b-a10b & 29 & 29 & 15.45 & 16.04 & 19 & 2.90 & 2.17 & 19 & 2.90 & 2.17 & 16 & 2.0625 \\
qwen-2.5-72b & 52 & 52 & 20.23 & 21.12 & 41 & 4.35 & 1.98 & 41 & 0.00 & 0.00 & 22 & 3.8182 \\
\rowcolor{gray!8}
llama-3.1-70b & 32 & 32 & 8.38 & 10.29 & 27 & 0.00 & 0.00 & 27 & 0.00 & 0.00 & 13 & 4.8462 \\
qwen-3.6-35b-a3b & 9 & 9 & 6.51 & 6.66 & 3 & 0.72 & 0.10 & 3 & 0.00 & 0.00 & 6 & \textbf{0.0000} \\
\rowcolor{gray!8}
qwen-3.6-coder-next & 15 & 15 & 9.05 & 10.06 & 6 & 0.00 & 0.00 & 6 & 0.00 & 0.00 & 10 & 1.2000 \\
llama-3.1-8b & 2 & 2 & 0.47 & 0.94 & 1 & 0.00 & 0.00 & 1 & 0.00 & 0.00 & 1 & 1.0000 \\
\bottomrule
\end{tabular}%
}

\endgroup
\end{table*}

\section{Engineering Knowledge Tables for Manufacturability}
\label{Engineering_Knowledge}
To systematically evaluate the manufacturability of a design, it is essential to consider the interplay between materials, fabrication processes, and assembly techniques. This section outlines the foundational engineering guidelines required to optimize designs for real-world production. Table \ref{tab:material_selection} presents a comprehensive overview of various materials, detailing their mechanical properties, process compatibility, and typical applications. Table \ref{tab:manufacturing_methods} delineates the technical capabilities, geometric constraints, and cost implications of common manufacturing methods, highlighting the trade-offs between precision and production feasibility. Finally, Table \ref{tab:connection_methods} categorizes standard mechanical connection methods, illustrating how different joint types constrain degrees of freedom (DoF) during assembly. Together, these references establish a structured framework for design for manufacturability (DFM).
\begin{table}[htbp]
    \centering
    \small
    \caption{Material Selection: Features and Typical Applications}
    \label{tab:material_selection}
    \begin{tabularx}{\textwidth}{l l X X}
        \toprule
        \textbf{Material} & \textbf{Compatible Processes} & \textbf{Features} & \textbf{Typical Applications} \\
        \midrule
        Timber & CNC, Laser Cutting & Easy to machine, medium strength, natural texture & Furniture, structural parts, DIY decor \\
        ABS & 3D Printing & High toughness, heat resistant & Structural components, enclosures \\
        PLA & 3D Printing & Easy to print, eco-friendly, low strength & Rapid prototypes, aesthetic models \\
        TPU & 3D Printing & Flexible, high elasticity & Seals, gaskets, cushioning parts \\
        Acrylic & CNC, Laser Cutting & Transparent, rigid but brittle & Housings, lighting, displays \\
        Resin & 3D Printing & High precision, fine detail, brittle & Intricate models, dental, figurines \\
        Sheet Metal & CNC, Laser Cutting & Lightweight, rapid forming & Chassis, enclosures \\
        Aluminum & CNC Milling & High strength-to-weight ratio, corrosion resistant & Structural parts, high-end housings \\
        Steel & CNC Milling & High strength and stiffness, heavy & Mechanical structures, supports \\
        \bottomrule
    \end{tabularx}
\end{table}

\begin{table}[htbp]
    \centering
    \scriptsize 
    \caption{Technical Specifications and Constraints of Manufacturing Methods}
    \label{tab:manufacturing_methods}
    \begin{tabularx}{\textwidth}{l X l l l X l}
        \toprule
        \textbf{Method} & \textbf{Constraints} & \textbf{Precision} & \textbf{Cost} & \textbf{Materials} & \textbf{Features} & \textbf{Diff.} \\
        \midrule
        CNC Milling & 2.5D/3-axis limits, tool radius offsets, max edge $<$ 2000 mm & $\pm$0.05-0.10 mm & Med-High & Wood, Al, Plastic & High precision; complex surfaces; restricted by tool paths & Low \\ \addlinespace
        Laser Cutting & 2D only, thickness $<$ 5 mm, max edge $<$ 2000 mm & $\pm$0.05-0.20 mm & Low-Med & Wood, Acrylic, Metal & Fast; low cost; ideal for interlocking assembly structures & Low \\ \addlinespace
        3D Printing & Build volume within 300 $\times$ 300 $\times$ 300 mm & $\pm$0.10-0.50 mm & Low & PLA, ABS, TPU & High freedom; rough surface, anisotropic strength, visible layer lines & Low \\ \addlinespace
        Injection Molding & Requires molds and draft angles; must be demoldable & $\pm$0.01-0.05 mm & High & Plastics & High precision and consistency; best for mass production & High \\ \addlinespace
        Silicone Casting & Dependent on master mold; shrinkage and bubble risks & $\pm$0.1-0.3 mm & Low-Med & Resin, Silicone & Suitable for small batch replication; cheaper than injection & High \\ \addlinespace
        Modular Assembly & Dependent on standard part sizes; tolerance stack-up & $\pm$0.10-0.50 mm & Low & Standard Parts, Profiles & Rapid construction; requires no additional manufacturing & Low \\
        \bottomrule
    \end{tabularx}
\end{table}
\begin{table}[htbp]
    \centering
    \small
    \caption{Classification and Features of Connection Methods}
    \label{tab:connection_methods}
    \begin{tabularx}{\textwidth}{l X m{3cm}} 
        \toprule
        \textbf{Joint Type} & \textbf{Features} & \textbf{Illustration} \\
        \midrule
        Interlocking & Typically constrains all 6 DoF when properly fitted & 
        \includegraphics[width=2.8cm, align=c]{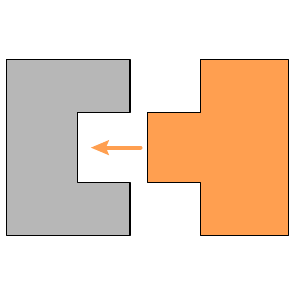} \\ \addlinespace
        
        Snap-fit & Nearly fully constrained; relies on elastic deformation & 
        \includegraphics[width=2.8cm, align=c]{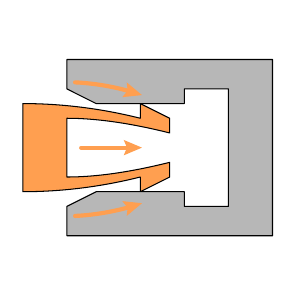} \\ \addlinespace
        
        Nailing & Partially constrains motion; allows rotation and axial slip (depending on friction) & 
        \includegraphics[width=2.8cm, align=c]{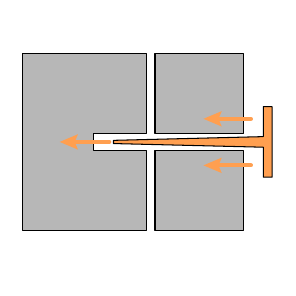} \\ \addlinespace
        
        Pivot & Constraints 5 DoF, allowing 1 rotational axis & 
        \includegraphics[width=2.8cm, align=c]{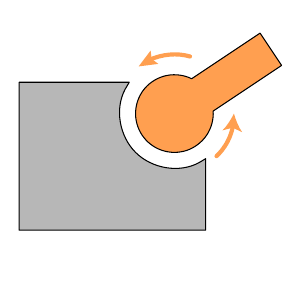} \\ \addlinespace
        
        Bonding & Fully constrained in all directions (permanent) & 
        \includegraphics[width=2.8cm, align=c]{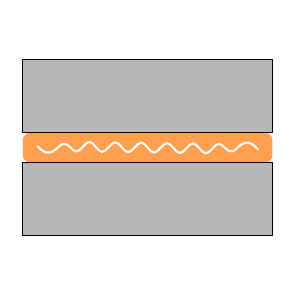} \\
        \bottomrule
    \end{tabularx}
\end{table}

\input{table/rubric}

\begin{figure*}[t]
    \centering
    \includegraphics[width=1.0\textwidth]{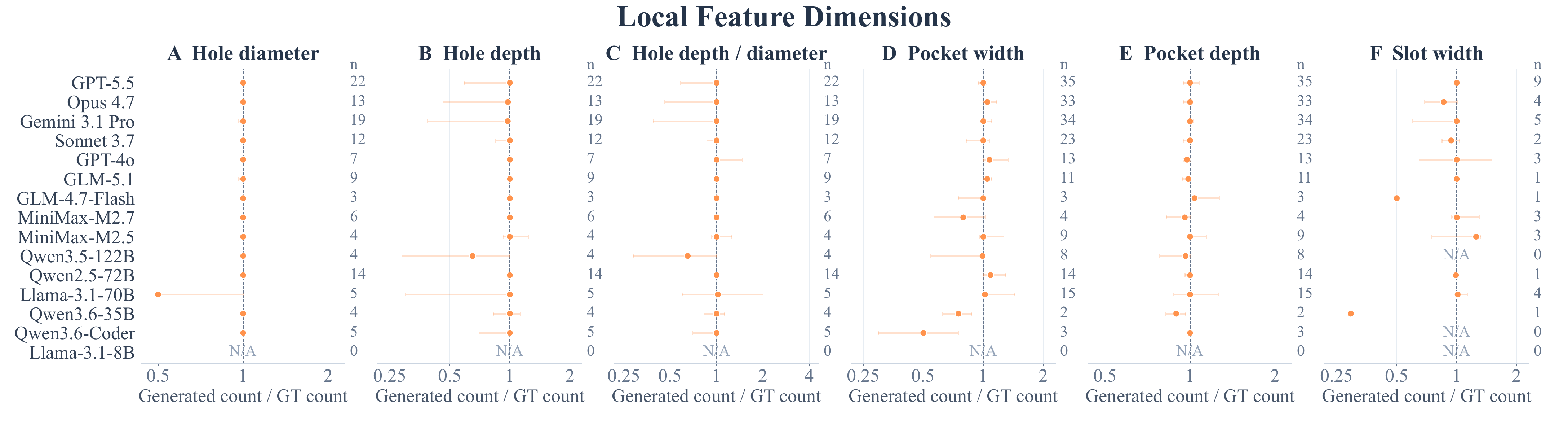}
    \caption{\textbf{Local feature dimensions.}
    Dots and light bars show the median and 25th--75th percentiles
    of per-design generated-to-GT dimension ratios.
    Each ratio compares within-design median descriptors,
    without individual feature matching.
    Only positive measured pairs are included.
    Ratios are uncapped; logarithmic axis ranges vary by panel.
    The dashed line marks 1; $n$ denotes eligible designs.}
    \label{fig:local-feature-dimensions}
\end{figure*}

\section{Prompt Templates}
\label{app:prompt_templates}

This appendix provides the prompt templates used throughout our benchmark construction and evaluation pipeline. Specifically, the first prompt is used in the human--LLM collaborative dataset construction stage to convert CAD assets into structured Design Specifications. The second prompt is used for design-specific evaluation, where a VLM generates task-specific rubrics for assessing design-intent alignment in terms of functionality, manufacturability, and assemblability.

\input{chapter/prompts/CAD_Design_Specification_Generator}

\input{chapter/prompts/Vision-LLM_Rubric_Generator}

\begin{table}[ht]
\centering
\caption{Per-model results across the three evaluation stages, judged by \textit{GPT-5.5}. The table reports code execution, geometric validity, and design-intent alignment scores under our funnel-style evaluation protocol. Bold marks the best score per block.}
\label{tab:overall-metrics_gpt5.5_}
\renewcommand{\arraystretch}{1.20}
\setlength{\tabcolsep}{3pt}
\scalebox{0.66}{
\begin{tabular}{l *{10}{S[table-format=2.2]}}
\toprule
 & {\textbf{Code Check (\%)}} & \multicolumn{5}{c}{\textbf{Geometry Check (\%)}} & \multicolumn{4}{c}{\textbf{Design Intent Alignment (\%)}} \\
\cmidrule(lr){2-2} \cmidrule(lr){3-7} \cmidrule(lr){8-11}
\textbf{Model} & {\footnotesize Sandbox Success} & {\footnotesize Watertight} & {\footnotesize Manifold} & {\footnotesize Self-Int.\ Free} & {\footnotesize Overlap Free} & {\footnotesize \textit{Geom.\ Valid}} & {\footnotesize \textbf{Functionality}} & {\footnotesize \textbf{Manufacturability}} & {\footnotesize \textbf{Assemblability}} & {\footnotesize \textit{Final Score}} \\
\midrule
\addlinespace[2pt]
\rowcolor{orange!10}
\multicolumn{11}{c}{\textit{\textbf{Closed-Source Models}}} \\
\addlinespace[1pt]
claude-opus-4.7 & 76.42 & \textbf{76.42} & \textbf{76.42} & \textbf{73.58} & 60.38 & 58.49 & 53.77 & 54.72 & 51.89 & 53.46 \\
\rowcolor{gray!7} claude-3.7-sonnet & 46.23 & 46.23 & 46.23 & 46.23 & 23.58 & 23.58 & 17.45 & 19.34 & 16.98 & 17.92 \\
gemini-3.1-pro & 65.09 & 59.43 & 63.21 & 63.21 & 62.26 & 58.49 & 53.77 & 55.19 & 52.83 & 53.93 \\
\rowcolor{gray!7} gpt-5.5 & \textbf{77.36} & 72.64 & 74.53 & \textbf{73.58} & \textbf{70.75} & \textbf{68.87} & \textbf{66.51} & \textbf{66.98} & \textbf{67.92} & \textbf{67.14} \\
gpt-4o & 36.79 & 36.79 & 36.79 & 36.79 & 13.21 & 13.21 & 8.02 & 9.91 & 8.49 & 8.81 \\
\addlinespace[1pt]
\rowcolor{gray!18} \textit{\textbf{Average}} & 60.38 & 58.30 & 59.43 & 58.68 & 46.04 & 44.53 & 39.91 & 41.23 & 39.62 & 40.25 \\
\midrule
\addlinespace[2pt]
\rowcolor{orange!10}
\multicolumn{11}{c}{\textit{\textbf{Open-Source Models}}} \\
\addlinespace[1pt]
glm-5.1 & 31.13 & 31.13 & 31.13 & 31.13 & \textbf{27.36} & \textbf{27.36} & \textbf{22.64} & \textbf{25.00} & \textbf{24.53} & \textbf{24.06} \\
\rowcolor{gray!7} glm-4.7-flash & 8.49 & 7.55 & 7.55 & 7.55 & 5.66 & 5.66 & 2.83 & 2.83 & 2.83 & 2.83 \\
minimax-m2.7 & 18.87 & 18.87 & 18.87 & 18.87 & 10.38 & 10.38 & 8.02 & 8.49 & 8.02 & 8.18 \\
\rowcolor{gray!7} minimax-m2.5 & 25.47 & 22.64 & 22.64 & 22.64 & 10.38 & 10.38 & 4.72 & 6.60 & 4.72 & 5.35 \\
qwen-3.5-122b-a10b & 27.36 & 25.47 & 25.47 & 25.47 & 13.21 & 13.21 & 10.38 & 10.38 & 10.38 & 10.38 \\
\rowcolor{gray!7} qwen-2.5-72b & \textbf{50.00} & \textbf{49.06} & \textbf{49.06} & \textbf{49.06} & 19.81 & 20.75 & 8.02 & 14.15 & 8.96 & 10.38 \\
llama-3.1-70b & 31.13 & 30.19 & 30.19 & 30.19 & 22.64 & 23.58 & 3.77 & 8.96 & 3.77 & 5.50 \\
\rowcolor{gray!7} qwen-3.6-35b-a3b & 8.49 & 8.49 & 8.49 & 8.49 & 5.66 & 5.66 & 4.72 & 4.72 & 4.72 & 4.72 \\
qwen-3.6-coder & 14.15 & 14.15 & 14.15 & 14.15 & 10.38 & 10.38 & 4.72 & 6.13 & 5.19 & 5.35 \\
\rowcolor{gray!7} llama-3.1-8b & 2.83 & 1.89 & 1.89 & 1.89 & 0.94 & 1.89 & 0.00 & 0.00 & 0.00 & 0.00 \\
\addlinespace[1pt]
\rowcolor{gray!18} \textit{\textbf{Average}} & 21.79 & 20.94 & 20.94 & 20.94 & 12.64 & 12.92 & 6.98 & 8.73 & 7.31 & 7.67 \\
\bottomrule
\end{tabular}
}
\end{table}

\section{Local Geometric Descriptors}
\label{app:local-geometric-analysis}
We organize the descriptors into local feature counts and local feature
dimensions, corresponding to Figures~\ref{fig:local-feature-counts}
and~\ref{fig:local-feature-dimensions}.
Table~\ref{tab:geometric-descriptor-hierarchy} defines each descriptor and
its interpretation. These are automatically extracted geometric summaries,
not matched manufacturing operations or validated manufacturability scores.

\begin{table*}[t]
\centering
\caption{Hierarchy, definitions, and interpretation of local geometric descriptors.
Dimension descriptors are summarized by their median within each design.
Lengths are in millimeters; counts and depth-to-diameter ratios are dimensionless.}
\label{tab:geometric-descriptor-hierarchy}
\begingroup\small
\setlength{\tabcolsep}{4pt}
\renewcommand{\arraystretch}{1.15}
\begin{tabularx}{\textwidth}{@{}p{.12\textwidth}p{.16\textwidth}>{\raggedright\arraybackslash}X>{\raggedright\arraybackslash}X@{}}
\toprule
Family & Descriptor & Definition / extraction & Interpretation relative to GT \\
\midrule
\multicolumn{4}{@{}l}{\textbf{I. Local feature counts}}\\
\addlinespace[2pt]
Holes & Hole count & Concave cylindrical faces with angular span $>1.6\pi$. & Fewer or more detected hole-like faces.\\
Fillets & Fillet count & Other concave cylindrical faces plus toroidal faces. & Fewer or more fillet-like patches, not verified rounding operations.\\
Pockets & Pocket count & Non-circular inner loops on planar faces with estimated length/width $\leq3$. & Fewer or more pocket-like openings, not necessarily distinct cavities.\\
Slots & Slot count & Non-circular inner loops on planar faces with estimated length/width $>3$. & Fewer or more elongated openings; their connection role is not assessed.\\
\midrule
\multicolumn{4}{@{}l}{\textbf{II. Local feature dimensions}}\\
\addlinespace[2pt]
Holes & Diameter & Median diameter of detected hole-like cylindrical faces. & Smaller or larger characteristic diameter.\\
 & Depth & Median axial extent of detected hole-like cylindrical faces. & Shorter or longer detected axial extent, not matched physical-hole depth.\\
 & Depth/diameter & Median of the individual detected faces' depth-to-diameter ratios. & Different hole proportions; depth, diameter, or the detected feature set may change.\\
Pockets & Width & Median estimated shorter in-plane extent of pocket-like loops. & Narrower or wider detected openings.\\
 & Depth & Median estimated adjacent-wall extent along the dominant face-normal axis. & Smaller or larger estimated recess depth.\\
Slots & Width & Median estimated shorter in-plane extent of slot-like loops. & Narrower or wider elongated openings.\\
\bottomrule
\end{tabularx}
\endgroup
\end{table*}

\paragraph{Extraction conventions.}
Cylindrical-face concavity is determined from the oriented surface normal
relative to the radial direction. Inner loops composed entirely of circular
arcs are excluded from pocket/slot detection. Loop dimensions are estimated
from axis-aligned bounding boxes after discarding the dominant face-normal
axis; adjacent-wall depth uses the corresponding axis. These approximations
can depend on orientation. Counts depend on the B-Rep representation, and
one physical feature can contribute multiple faces or openings.

\paragraph{Normalization and aggregation.}
We pair generated and GT records by design identity, requiring successful
extraction and at least one solid in each imported shape. This does not
imply passing all benchmark geometric-validity checks.
Let $x^{\mathrm{gen}}_{i,m,k}$ and $x^{\mathrm{GT}}_{i,k}$ denote descriptor
$k$ for design $i$, generated by model $m$ and in GT, respectively.
For feature counts, GT counts must be positive; generated zero counts are
retained, and the per-design ratio is
\begin{equation}
r_{i,m,k}=\min\!\left(x^{\mathrm{gen}}_{i,m,k}/x^{\mathrm{GT}}_{i,k},\,2\right).
\end{equation}
For dimensions, both values must be positive, and the ratio is uncapped.
Missing or unmeasurable dimensions are excluded rather than assigned zero.
Dimension ratios compare within-design medians without matching individual
features; the median depth-to-diameter ratio is not necessarily the ratio
of median depth to median diameter.
Across eligible designs, dots show the median ratio and light bars show
its 25th--75th percentiles, computed by linear interpolation. These are
interquartile ranges, not confidence intervals. We report eligible sample
sizes $n$, which may differ across models and descriptors.
Count panels use linear axes, whereas dimension panels use logarithmic
axes with panel-specific ranges. A collapsed interval does not imply that
all designs agree. Numerical agreement does not establish correct feature
placement, specification compliance, or manufacturability.

%% file: chapter/evaluation_details.tex
\section{Implementation Details for Structural and Interface Evaluation}
\label{app:evaluation_details}

This appendix provides the complete notation and procedures used by the
structural and interface checks in Section~\ref{sec:evaluation}. The main
paper summarizes the evaluation logic and metrics, while the details below
support reproducibility.

\subsection{Evaluation Notation}

\begin{table*}[t]
\centering
\small
\caption{Notation used in the evaluation protocol.}
\label{tab:evaluation_notation}
\begin{tabular}{p{0.16\textwidth} p{0.76\textwidth}}
\toprule
\textbf{Symbol} & \textbf{Definition} \\
\midrule

$\mathcal{S}$
& Design Specification describing the target CAD object, its components,
interfaces, dimensions, and functional requirements. \\

$C$
& Generated CAD model represented by the exported STEP B-Rep. \\

$S_i$
& The $i$-th solid component in the generated STEP model. \\

$\mathcal{G}^{*}=(\mathcal{V}^{*},\mathcal{E}^{*})$
& Ground-truth Physical Assembly Graph. \\

$\widehat{\mathcal{G}}
=(\widehat{\mathcal{V}},\widehat{\mathcal{E}})$
& Physical Assembly Graph recovered from the generated CAD model. \\

$v_i$
& A component vertex in a Physical Assembly Graph. \\

$e_{ij}$
& A physical interface between components $v_i$ and $v_j$. \\

$\ell_i$
& Semantic component label associated with vertex $v_i$. \\

$\tau_{ij}$
& Physical interface type associated with edge $e_{ij}$,
e.g., a rectangular mortise-and-tenon connection. \\

$\mathbf{p}_i$
& Continuous design-parameter vector associated with component $v_i$,
such as width, height, and thickness. \\

$\Omega$
& Valid Parameter Space containing admissible ranges for required
continuous component and interface parameters. \\

$e$
& Index of a localized fitting interface. \\

$i$
& Index of a side wall within a rectangular fitting interface,
$i\in\{1,\ldots,4\}$. \\

$k$
& Index of one of the two orthogonal cross-sectional directions,
$k\in\{1,2\}$. \\

$\mathbf{c}^{m}_{ei}$
& Area centroid of the $i$-th male-side face of interface $e$. \\

$\mathbf{c}^{r}_{ei}$
& Area centroid of the corresponding receiver-side face. \\

$\mathbf{n}^{m}_{ei}$
& Outward unit normal of the male-side face. \\

$g_{ei}$
& Single-side clearance of the $i$-th paired side faces:
$(\mathbf{c}^{r}_{ei}-\mathbf{c}^{m}_{ei})^{\top}
\mathbf{n}^{m}_{ei}$. \\

$w^{m}_{ek}$
& Male-component width of interface $e$ along cross-sectional
direction $k$. \\

$w^{r}_{ek}$
& Receiver width of interface $e$ along cross-sectional direction $k$. \\

$\Delta_{ek}$
& Total fitting clearance along direction $k$,
defined as $w^{r}_{ek}-w^{m}_{ek}$ and equivalently as the sum of the
two corresponding single-side clearances. \\

$\Delta^{*}_{ek}$
& Ground-truth total clearance for interface $e$ along direction $k$. \\

$[L_{ek},U_{ek}]$
& Task-specified admissible interval for the total clearance
$\Delta_{ek}$. \\

$\mathcal{Q}$
& Set of interface--direction pairs with both valid predicted and
ground-truth clearance measurements. \\

$N_{\Delta}$
& Number of evaluated clearance measurements,
$N_{\Delta}=|\mathcal{Q}|$. \\

$\operatorname{GapMAE}$
& Mean absolute error between predicted and ground-truth total
clearances. \\

$\operatorname{GED}$
& Graph Edit Distance between the recovered and ground-truth
Physical Assembly Graphs. \\

\bottomrule
\end{tabular}
\end{table*}

\clearpage

\begin{algorithm}[H]
\caption{Reference-Guided Component Name Correspondence}
\label{alg:naming}

\begin{algorithmic}

\Require Generated solids $G=\{S_i\}_{i=1}^{n}$,
labeled reference solids
$R=\{(T_j,\ell_j)\}_{j=1}^{m}$,
unmatched cost $\tau$, ambiguity margin $\delta$

\Ensure Suggested component names $M$

\vspace{0.5em}

\Statex
\textcolor{blue}{\texttt{// Step 1: Compute Relative Geometric Descriptors}}

\State Compute normalized descriptors
$\{(\mathbf{p}_i^G,\mathbf{d}_i^G,f_i^G)\}_{i=1}^{n}$
for generated model $G$

\State Compute normalized descriptors
$\{(\mathbf{p}_j^R,\mathbf{d}_j^R,f_j^R)\}_{j=1}^{m}$
for reference model $R$

\vspace{0.5em}

\Statex
\textcolor{blue}{\texttt{// Step 2: Compute Pairwise Matching Costs}}

\For{each generated solid $S_i$}
    \For{each reference solid $T_j$}

        \State Compute
        \[
        D_{ij}
        =
        w_p\|\mathbf{p}_i^G-\mathbf{p}_j^R\|_2
        +
        \frac{w_d}{3}
        \sum_{k\in\{x,y,z\}}
        \left|
        \log\frac{d_{ik}^G}{d_{jk}^R}
        \right|
        +
        w_f
        \left|
        \log\frac{f_i^G}{f_j^R}
        \right|
        \]

    \EndFor
\EndFor

\vspace{0.5em}

\Statex
\textcolor{blue}{\texttt{// Step 3: Global Assignment with Unmatched Options}}

\State Construct augmented cost matrix
\[
\widetilde{D}
=
\left[
D
\mid
\tau\mathbf{1}_{n\times n}
\right]
\]

\State Compute global assignment
\[
\pi^\star
=
\operatorname{Hungarian}(\widetilde{D})
\]

\vspace{0.5em}

\Statex
\textcolor{blue}{\texttt{// Step 4: Reject Ambiguous Matches and Transfer Names}}

\State Initialize $M\gets[\,]$

\For{each generated solid $S_i$}

    \State $j\gets\pi^\star(i)$

    \State
    $\widehat{j}_i\gets\bot$,
    $\widehat{\ell}_i\gets\bot$

    \State
    $A_i\gets\TopThree(D_{i,:})$

    \State
    $r_i\gets\Gap(D_{i,:})$

    \If{$j\leq m$}

        \State
        $q_i\gets\Gap(D_{:,j})$

        \If{
        $j=\arg\min_t D_{it}$
        \textbf{ and }
        $i=\arg\min_t D_{tj}$
        }

            \If{
            $D_{ij}<\tau$
            \textbf{ and }
            $r_i\geq\delta$
            \textbf{ and }
            $q_i\geq\delta$
            }

                \If{$\ValidName(\ell_j)$}

                    \State
                    $\widehat{j}_i\gets j$

                    \State
                    $\widehat{\ell}_i\gets\ell_j$

                \EndIf
            \EndIf
        \EndIf
    \EndIf

    \State Append
    $(\operatorname{id}(S_i),
    \widehat{j}_i,
    \widehat{\ell}_i,
    A_i,
    \mathrm{confirmed}=\mathrm{false})$
    to $M$

\EndFor

\vspace{0.4em}

\State \Return $M$

\end{algorithmic}
\end{algorithm}

\begin{algorithm}[H]
\caption{Rectangular Mortise-and-Tenon Detection
from Local Four-Wall Evidence}
\label{alg:mortise-tenon}
\footnotesize

\begin{algorithmic}[1]

\Require Generated STEP model $C$;
$\tau_g=3\,\mathrm{mm}$,
$\tau_\theta=1^\circ$,
$\tau_A=0.5\,\mathrm{mm}^2$,
$\tau_\rho=0.05$,
$\tau_L=0.5\,\mathrm{mm}$;
search budget $T_{\max}=200{,}000$

\Ensure Hypothesis set $\mathcal H$ and diagnostic records $\mathcal U$

\Statex \textbf{Step 1: Extract geometry and initialize}

\State Read solids $\{S_i\}$ and planar-face sets
$\{\mathcal F_i^{\mathrm{pl}}\}$;
retain original solid and face IDs

\State
$\mathcal P\gets\varnothing$;
$\mathcal H\gets\varnothing$;
$\mathcal U\gets\varnothing$;
$t\gets0$

\Statex \textbf{Step 2: Build cross-solid planar-pair records}

\For{each solid-index pair $(i,j)$ with $i<j$}

    \If{solid bounding-box distance $>\tau_g$}
        \State \textbf{continue}
    \EndIf

    \For{each $(f_a,f_b)\in
          \mathcal F_i^{\mathrm{pl}}\times
          \mathcal F_j^{\mathrm{pl}}$}

        \If{face bounding-box distance $>\tau_g$}
            \State \textbf{continue}
        \EndIf

        \State
        $\alpha\gets
        \arccos(
        \operatorname{clip}
        (-\mathbf n_a^\top\mathbf n_b,-1,1))$

        \State
        $g\gets
        (\mathbf c_b-\mathbf c_a)^\top\mathbf n_a$

        \If{$\alpha>\tau_\theta$ or $|g|>\tau_g$}
            \State \textbf{continue}
        \EndIf

        \State
        $\widetilde f_b\gets
        \Call{AlignFace}{f_a,f_b}$

        \State
        $A\gets
        \operatorname{Area}
        (f_a\cap\widetilde f_b)$

        \State On geometric failure,
        append a diagnostic to $\mathcal U$
        and \textbf{continue}

        \If{$A\geq\tau_A$ and
            $A/\min(A_a,A_b)\geq\tau_\rho$}

            \State
            $\mathcal P\gets
            \mathcal P\cup
            \{(i,j,f_a,f_b,g,A)\}$

        \EndIf

    \EndFor
\EndFor

\Statex \textbf{Step 3: Construct local tenons and match receiver walls}

\State
$\mathcal D\gets
\{(i,j),(j,i)
\mid
(i,j,f_a,f_b,g,A)\in\mathcal P\}$
\Comment{Both role assignments}

\For{each $(i,j)\in\mathcal D$}

    \State Treat $S_i$ as the proposed male
    and $S_j$ as the receiver

    \State
    $\mathcal P_{ij}\gets
    \{p\in\mathcal P
    \mid
    p\text{ connects solids }S_i\text{ and }S_j\}$

    \State
    $\mathcal M_{ij}\gets
    \{f\in\mathcal F_i^{\mathrm{pl}}
    \mid
    f\text{ occurs in }\mathcal P_{ij}\}$

    \For{each combination $Q$ of two convex opposing-wall pairs
          from $\mathcal M_{ij}$}

        \State $t\gets t+1$

        \If{$t>T_{\max}$}
            \State Append a search-truncation diagnostic to $\mathcal U$
            \State \Return $(\mathcal H,\mathcal U)$
        \EndIf

        \If{$Q$ lacks four distinct walls,
            orthogonal pair directions,
            or vertex data}
            \State \textbf{continue}
        \EndIf

        \State Compute local axes
        $(\mathbf x,\mathbf y,\mathbf z)$,
        center $\mathbf o$,
        widths $w,d$,
        and common axial length $L$

        \If{$L<\tau_L$
            or the walls do not cover the same local section}
            \State \textbf{continue}
        \EndIf

        \State
        $\mathcal X\gets
        \{
        \mathbf o+uw\mathbf x+vd\mathbf y
        \mid
        u,v\in\{-\tfrac14,0,\tfrac14\}
        \}$

        \If{$\mathcal X\not\subset S_i$
            or $\mathbf o\in S_j$}
            \State \textbf{continue}
        \EndIf

        \For{each male wall $f\in Q$}

            \State
            $\mathcal R_f\gets
            \{
            r\in\mathcal F_j^{\mathrm{pl}}
            \mid
            (f,r)\text{ is represented in }\mathcal P_{ij},
            \;
            (\mathbf c_r-\mathbf o)^\top\mathbf n_r
            <-10^{-5}\,\mathrm{mm}
            \}$

            \If{$\mathcal R_f\neq\varnothing$}

                \State
                $r^*(f)\gets
                \operatorname*{arg\,max}_{r\in\mathcal R_f}
                A_{fr}$
                \Comment{Overlap from the pair record}

            \EndIf

        \EndFor

        \If{$\mathcal R_f\neq\varnothing$
            for every $f\in Q$}

            \State Construct $h$ with solid/face IDs,
            pair evidence, local coordinates,
            section dimensions, and recorded gaps

            \State Insert $h$ into $\mathcal H$
            unless its local geometric key already exists

        \EndIf

    \EndFor
\EndFor

\Statex \textbf{Step 4: Return hypotheses and diagnostics}

\State \Return $(\mathcal H,\mathcal U)$

\end{algorithmic}
\end{algorithm}


%% file: table/rubric.tex
\begin{table}[htbp]
\centering
\caption{Evaluation criteria used in the design-intent alignment stage.}
\label{tab:design_intent_criteria}
\renewcommand{\arraystretch}{1.25}
\small
\begin{tabularx}{\textwidth}{@{} 
    >{\raggedright\arraybackslash}p{0.16\textwidth}
    >{\raggedright\arraybackslash}p{0.18\textwidth}
    >{\raggedright\arraybackslash}X
    >{\raggedright\arraybackslash}p{0.28\textwidth}
    @{}}
\toprule
\textbf{Criterion} & \textbf{Sub-criterion} & \textbf{Evaluation Question} & \textbf{Evidence Used} \\ \midrule

\multirow{2}{*}{\textbf{Functionality}}
& Functional
& Can the object support its intended primary use and auxiliary function?
& Design Specification; functional requirements; visual proportions and structures in \texttt{<Generated\_SVG>}. \\ \cmidrule(l){2-4}

& Robust
& Can the object remain stable and structurally reliable during use?
& Mechanical condition; component topology; joint relationships; support posture and load-transfer path in \texttt{<Generated\_SVG>}. \\ 




\bottomrule

\end{tabularx}
\end{table}

%% file: chapter/prompts/CAD_Design_Specification_Generator.tex
\begin{promptbox}[System Prompt: CAD Design Specification Generator]
System Prompt: CAD Design Specification Agent

## Role
You are an Expert CAD Systems Engineer and Manufacturing Specialist. Your goal is to analyze provided CAD data (CadQuery code, SVG/STP files) and decompose it into a formal, engineering-grade **Design Specification** (Design Task Document) in **English**.

## Core Task
Generate a structured document that defines the design intent, physical constraints, and manufacturing requirements of a 3D model based on its B-Rep logic and parametric script.

## Constraints for Technical Selection
You MUST select the **Material**, **Manufacturing Method**, and **Connection Method** strictly from the provided technical tables in the `Reference Tables` section below. Do not hallucinate external methods.

## Inference Logic
* **Analyze Joint Type:** Look at boolean operations in CadQuery (`cut`, `fuse`, `intersect`). If a component has a protruding box (`tenon`) and another has a subtracted hole (`socket/mortise`), select **Mortise & Tenon**.
* **Analyze Manufacturing:** Based on the geometry (e.g., 2.5D shapes are suitable for CNC/Laser, complex 3D shapes for Printing) and the selected material.
* **Analyze Parameters:** Extract variable names and ranges from dictionaries like `PARAM_RANGES` or script constants.

## Output Format Requirements
Your response MUST strictly follow the Markdown structure below:

# Design Specification

## Design Goal
[Concise description of what the object is and its primary function.]

## Geometry and Dimensions
Approx. [Width] mm  [Depth] mm  [Height] mm.

## Material
[Select strictly from Table 1]

## Manufacturing Method
[Select strictly from Table 2]

## Connection Method (Joint Type)
[Select strictly from Table 3]

## Mechanical Condition
[User scenario, e.g., Single-person seating, load-bearing storage, etc.]

## Structural Features
[List main components separated by semicolons; e.g., Seat panel; four legs; backrest.]

## Special Requirements
[Any constraints like "Keep assembly split unchanged".]

## Planned Component Quantity
[Count of independent solid bodies]

## Component Names
- [Part Name 1]
- [Part Name 2]
...

## Adjustable Parameters
- **[Parameter Name]**: [Value] ([Min] ~ [Max] mm). [Reasoning/Constraint description].

## Component Details

### [Component Number]. [Component Name]
[Brief description of the part's role in the assembly.]
* **Component Purpose**: [Specific functional role].
* **Assembly Direction**: [Vector/Direction, e.g., Vertical insertion along +Z axis].
* **Connection & Kinematics**: [Joint Type from Table 3] ([Degrees of Freedom/Kinematics description from Table 3]).

---

## Component Assembly Graph (Textual)
[Part A] -> [Part B] | Joint: [Type] | Note: [Description of the interface]

---
## Reference Tables (Knowledge Base)
...

\end{promptbox}

%% file: chapter/prompts/Vision-LLM_Rubric_Generator.tex
\begin{promptbox}[System Prompt: Vision-LLM Rubric Generator]
# Role and Objective

You are a top-tier expert in CAD (Computer-Aided Design), physical manufacturing, and assembly evaluation. Your task is to dynamically write a rigorous 0--1 Pass/Fail evaluation rubric for an AI-generated 3D model.

The rubric you generate will be injected into the `<Evaluation_Rubric>` placeholder and used by a downstream Vision-LLM judge to evaluate `<Generated_SVG>`.

# Background Knowledge Base

When reasoning about tolerances, joint relationships, and manufacturability, you must cross-reference the following three tables.

## Table 1: Manufacturing Methods

| Method | Constraints | Precision / Tolerance | Features |
| :--- | :--- | :--- | :--- |
| CNC Milling | Limited by 2.5D / 3-axis machining; tool radius compensation must be considered | 0.05-0.10 mm | High precision; can machine complex surfaces; constrained by tool paths |
| Laser Cutting | 2D only; thickness < 5 mm | 0.05-0.20 mm | Fast; low cost; highly suitable for interlocking assembly structures |
| 3D Printing | Build volume limitations; overhangs may require support | 0.05-0.50 mm | High geometric freedom; supports complex internal cavities and topology; may show layer lines |
| Molding | Requires molds and draft angles; must be demoldable; may depend on standard part sizes | 0.01-0.50 mm | Relies on master molds or standard libraries; suitable for mass production; inaccessible closed dead cavities are forbidden |

## Table 2: Material Selection

| Material | Compatible Processes | Features |
| :--- | :--- | :--- |
| Timber | CNC, Laser Cutting | Easy to process; medium strength; natural texture |
| PLA-ABS | 3D Printing | PLA: easy to print, eco-friendly, low strength; ABS: tough and heat-resistant |
| Acrylic | CNC, Laser Cutting | Transparent, rigid, but brittle |
| Resin | 3D Printing / Molding | High precision, fine details, brittle |
| Aluminum | CNC Milling | High strength-to-weight ratio; corrosion-resistant |
| Steel | CNC Milling | High strength and stiffness; heavy |

## Table 3: Connection Methods

| Joint Type | Features |
| :--- | :--- |
| Interlocking | When properly assembled, it usually constrains all 6 DoFs |
| Snap-fit | Almost fully constrains motion; relies on elastic deformation |
| Nailing / Pinning | Partially constrains motion; may allow rotation and axial sliding depending on friction |
| Pivot / Hinge | Constrains 5 DoF and allows 1 rotational axis |
| Bonding | Fully constrains motion in all directions; permanent connection |

# Critical Constraint: Cross-modal Translation Principle

The downstream judge cannot see the source code or the original design document. It can only read your written rubric and inspect `<Generated_SVG>` and `<Reference_SVG>`.

Therefore, when writing the rubric, you must translate the information in `<Task_Doc>` and the background knowledge base into visual inspection instructions that the judge can follow.

Important terminology rule:

- Use **node** only when referring to components in the Component Assembly Graph.
- Use **joint** only when referring to physical connection mechanisms between components.
- Do not use "joint node" or confuse graph nodes with physical joints.
- Normalize joint names before writing the rubric: use **Interlocking** instead of Mortise & Tenon; use **Nailing / Pinning** instead of dowel joint.

# Required Rubric Generation Method

You must generate exactly the following six evaluation dimensions.

## 1. Assembly Readiness

You need to extract `[Component Assembly Graph (Textual)]` from `<Task_Doc>`.

In the rubric, instruct the judge to infer the Component Assembly Graph from `<Generated_SVG>` by treating each visible component as a graph node and each physical connection relationship as an edge. The inferred graph must then be compared with the target graph described in the task.

The Pass criterion must state that the visually inferred node-edge topology is semantically consistent with the target Component Assembly Graph.

The Fail criterion must state that the graph topology is incorrect, such as missing nodes, extra nodes, wrong node-to-node links, disconnected components, or an incorrect central hub.

## 2. Joint Design

You need to parse `<Task_Doc>` to extract the required joint type and assembly direction. You must also consult Table 3 to obtain the features of that joint type.

In the rubric, the Pass criterion must state that the macroscopic connection location and direction are correct, and that the visual joint design conforms to the required joint type and its physical constraints. You must explicitly mention the features obtained from Table 3, such as whether the joint fully constrains all directions, relies on elastic deformation, or allows a rotational axis.

The Fail criterion must state that the joint design contradicts the specified joint type or its constraint behavior, or that severe rigid-body interpenetration, illegal floating, or physically impossible attachment occurs.

## 3. Tolerance

You need to extract the manufacturing method from `<Task_Doc>` and consult Table 1 to obtain its precision / tolerance range.

In the rubric, the Pass criterion must explicitly mention the process-level precision value. It must also instruct the judge to use the seam, clearance, wall-thickness, or contact-gap appearance in `<Reference_SVG>` as the visual scale reference. `<Generated_SVG>` should show appropriate independent boundaries, seams, or clearances.

The Fail criterion must state that seams disappear entirely due to illegal fusion, or that the visible gaps are extremely exaggerated and would prevent real-world fitting, slicing, assembly, or functional use.

## 4. Functional Adaptation

You need to infer 1-2 core functions of the object and clearly separate them into:

- **Must-have function**, usually weighted around 70
- **Nice-to-have function**, usually weighted around 30

In the rubric, the Pass criterion must state that the visual proportions and structures fully satisfy the must-have function and include the basic structures needed for the nice-to-have function.

The Fail criterion must give concrete fatal-error examples, such as the must-have function being completely lost, or the nice-to-have feature becoming extremely distorted and damaging the overall must-have use.

## 5. Usage Stability

You need to extract the intended stability, anti-tipping, load-bearing, or support behavior from `<Task_Doc>`, and infer the force-transfer path based on the component topology and physical joints.

In the rubric, the Pass criterion must state that `<Generated_SVG>` shows a stable support posture, reasonable load transfer, and load-bearing members with safe visual thickness.

The Fail criterion must list cases that would inevitably cause tipping, collapse, or structural instability, such as centered support points, uneven legs, extremely thin load-bearing members, insufficient base contact, or disconnected load paths.

## 6. Manufacturability

You need to extract `[Material]` and `[Manufacturing Method]` from `<Task_Doc>`. You must cross-check Table 1 and Table 2 to verify the process constraints and material features.

In the rubric, the Pass criterion must state that `<Generated_SVG>` uses conventional geometries compatible with the specified process and the physical features of the material.

The Fail criterion must list topology or geometry features that violate the process or material constraints, such as inaccessible internal dead cavities for CNC machining, zero-thickness surfaces, non-manifold geometry for 3D printing, or extremely thin load-bearing cantilevers made from brittle materials.

# Input Modalities

Only you can see the following inputs. Do not reveal source code or raw design documents to the downstream judge.

1. `<Task_Doc>`: the design specification, including design goals, component list, parameter ranges, and assembly graph.
2. `<Reference_Code>`: the ground-truth CAD logic and spatial coordinates.
3. `<Reference_SVG>`: the visual anchor / reference image.

# Output Template

Strictly follow the Markdown template below.

You must use the exact terms `<Reference_SVG>` and `<Generated_SVG>` in the rubric.

Write the instructions as if you are directly guiding the downstream judge.

Do not include a separate "Core Focus" field. Instead, merge all necessary inspection guidance into the Pass criterion.

### 1. Assembly Readiness

* **Evaluation Rubric (0-1 Score)**:

    * **1 Point (Pass)**: [Instruct the judge to infer the Component Assembly Graph from `<Generated_SVG>`; describe the required target graph using node-edge language; state what counts as a correct visual topology.]

    * **0 Points (Fail)**: [Describe incorrect graph topology, such as missing nodes, extra nodes, incorrect links, disconnected components, or wrong hub structure.]

---

### 2. Joint Design

* **Evaluation Rubric (0-1 Score)**:

    * **1 Point (Pass)**: [Describe the required joint type, physical constraint behavior, connection location, assembly direction, and acceptable visual appearance.]

    * **0 Points (Fail)**: [Describe joint-type mismatch, wrong direction, missing attachment, illegal floating, severe interpenetration, or physically impossible connection.]

---

### 3. Tolerance

* **Evaluation Rubric (0-1 Score)**:

    * **1 Point (Pass)**: [Mention the manufacturing precision value; instruct the judge to use `<Reference_SVG>` as the visual scale reference; describe acceptable seams, clearances, wall thickness, or contact gaps.]

    * **0 Points (Fail)**: [Describe illegal fusion, exaggerated gaps, missing clearances, invalid wall thickness, or tolerance errors that make the object impossible to assemble, slice, or use.]

---

### 4. Functional Adaptation

* **Evaluation Rubric (0-1 Score)**:

    * **1 Point (Pass)**: [State the must-have function and nice-to-have function; describe how `<Generated_SVG>` visually satisfies them.]

    * **0 Points (Fail)**: [Give concrete fatal examples where the must-have function is lost, or the nice-to-have feature severely damages the must-have use.]

---

### 5. Usage Stability

* **Evaluation Rubric (0-1 Score)**:

    * **1 Point (Pass)**: [Describe the intended support / load-transfer / anti-tipping behavior; specify what stable visual geometry should look like.]

    * **0 Points (Fail)**: [Describe inevitable tipping, collapse, deformation, unstable support, missing support points, or broken load paths.]

---

### 6. Manufacturability

* **Evaluation Rubric (0-1 Score)**:

    * **1 Point (Pass)**: [State the material and manufacturing process; describe geometries compatible with the process and material.]

    * **0 Points (Fail)**: [Describe geometry or topology that violates material/process constraints, such as zero-thickness surfaces, inaccessible cavities, non-manifold geometry, impossible overhangs, or physically fragile structures.]

---

# Few-shot Reference

## Example 1: Four-legged Dining Chair with Backrest

### 1. Assembly Readiness

* **Evaluation Rubric (0-1 Score)**:

    * **1 Point (Pass)**: Ask the judge to infer the Component Assembly Graph from `<Generated_SVG>` by treating each visible component as a graph node and each physical connection relationship as an edge. The inferred graph must match the target topology: `seat_panel` is the only central hub node; `front_left_leg`, `front_right_leg`, `rear_left_leg`, and `rear_right_leg` each connect to the four lower corners of `seat_panel`; `backrest_panel` connects to the rear side of `seat_panel`. The object should be visually identifiable as a 6-component dining chair consisting of 1 seat panel, 4 legs, and 1 backrest.

    * **0 Points (Fail)**: The graph topology is incorrect. For example, the backrest is floating instead of connected to the seat panel; a leg is connected to the backrest or to another leg instead of the seat panel; any leg is missing; extra components such as armrests or crossbars alter the required 6-component structure; or the seat panel is no longer the central hub node.

---

### 2. Joint Design

* **Evaluation Rubric (0-1 Score)**:

    * **1 Point (Pass)**: Ask the judge to inspect the assembly regions shown in `<Reference_SVG>` and verify that the macroscopic connection locations and assembly directions in `<Generated_SVG>` are correct. The four legs should insert upward along the +Z direction into sockets on the underside of the seat panel, and the backrest should press downward along the -Z direction into the long rear slot of the seat panel. The required joint is Interlocking: when properly assembled, it usually constrains all 6 degrees of freedom. Visually, the leg protrusions should appear to enter the seat sockets, and the backrest strip should appear to enter the rear slot with tight engagement. Minor line overlap caused by SVG rendering or projection is acceptable.

    * **0 Points (Fail)**: The joint design contradicts the Interlocking behavior. For example, the legs merely touch the outer sides of the seat without an insertion relationship; the backrest floats above the seat; the slot or insertion direction is wrong; severe interpenetration occurs between rigid bodies; or the connection points are so misaligned that the protrusions could not enter the corresponding sockets or constrain all 6 degrees of freedom.

---

### 3. Tolerance

* **Evaluation Rubric (0-1 Score)**:

    * **1 Point (Pass)**: The manufacturing process is CNC Milling, whose expected precision is 0.05-0.10 mm. Ask the judge to use the thin visual seams and interlocking contact boundaries in `<Reference_SVG>` as the scale reference. In `<Generated_SVG>`, the boundaries between the seat panel and legs, and between the seat panel and backrest, should remain visible as very fine seams or tight contact lines. The fit should visually suggest a CNC-machined tight interlocking assembly: the clearance is small, but the component boundaries do not disappear.

    * **0 Points (Fail)**: The seams disappear completely, making the seat, legs, and backrest look illegally fused into a single body; or the visible gaps are extremely exaggerated, such as legs visibly separated from the seat or a large gap between the backrest and the rear slot, making real interlocking fitting or load transfer impossible.

---

### 4. Functional Adaptation

* **Evaluation Rubric (0-1 Score)**:

    * **1 Point (Pass)**: The must-have function, weighted around 70

    * **0 Points (Fail)**: The must-have function is completely lost. For example, the seat is extremely narrow or paper-thin and cannot visually support a person; the legs are so short or tall that the seat height is unusable; the backrest appears at the front or side of the chair or is missing; or the backrest is extremely distorted, heavily tilted, or passes through the center of the seat in a way that destroys the sitting function.

---

### 5. Usage Stability

* **Evaluation Rubric (0-1 Score)**:

    * **1 Point (Pass)**: Ask the judge to evaluate anti-tipping posture, support polygon size, and the visual load-transfer path. The intended force path is: human weight  `seat_panel`  four corner legs  ground. `<Generated_SVG>` should resemble `<Reference_SVG>` by showing four vertical legs distributed near the seat edges or corners, with consistent leg length so the seat can remain level. The legs and seat panel should have safe visual thickness. The center of gravity should plausibly fall inside the support region formed by the four legs. Although the backrest introduces rearward torque, it should be stabilized through the rear legs and the interlocking connection to the seat panel.

    * **0 Points (Fail)**: The object would inevitably tip or fail structurally. For example, the four legs are clustered near the center, producing a tiny support polygon; only two or three legs touch the ground; leg lengths are inconsistent; the rear legs are missing while the backrest is tall, causing backward tipping; or the legs or seat are visually as thin as lines and cannot support a seated person.

---

### 6. Manufacturability

* **Evaluation Rubric (0-1 Score)**:

    * **1 Point (Pass)**: The material is Timber and the manufacturing process is CNC Milling. Timber is compatible with CNC machining, easy to process, medium strength, and has natural texture. CNC Milling has 0.05-0.10 mm precision but is constrained by 2.5D / 3-axis tool paths and tool radius. `<Generated_SVG>` should use conventional wood-CNC geometries: a rectangular seat panel, four square-post legs, a vertical backrest panel, and understandable rectangular interlocking interfaces. The components should remain visually distinguishable as independent closed solids, and the original 6-component split should not be altered.

    * **0 Points (Fail)**: `<Generated_SVG>` shows geometry or topology that violates Timber or CNC Milling constraints. Examples include zero-thickness seat or backrest surfaces, inaccessible internal dead cavities that a CNC tool cannot reach, extremely thin load-bearing wooden cantilevers or needle-like legs, margins so small that wood splitting is visually likely, illegal fusion of multiple components into an inseparable body, or overly organic surfaces that cannot reasonably be produced by conventional 3-axis wood CNC machining.

---

## Example 2: Wave Vase

### 1. Assembly Readiness

* **Evaluation Rubric (0-1 Score)**:

    * **1 Point (Pass)**: Ask the judge to infer the Component Assembly Graph from `<Generated_SVG>`. The target graph is `wave_vase_body -> Standalone`, with Joint: None. The inferred topology must match this target: the model should contain exactly one standalone graph node, `wave_vase_body`, with no extra components, no separated parts, and no movable joints. The outer shell, inner cavity, base, and top rim should all belong to the same continuous vase body.

    * **0 Points (Fail)**: The graph topology is incorrect. For example, the model is split into multiple separated petal-like panels; the outer shell and inner liner appear as two independent components; the base is detached from the body; extra components such as handles, stands, or lids appear; or the object looks like an assembly of multiple parts rather than one standalone body.

---

### 2. Joint Design

* **Evaluation Rubric (0-1 Score)**:

    * **1 Point (Pass)**: The required joint type is N/A because this is a single continuous body with no physical assembly joint, no hinge, no bonding, no snap-fit, and no interlocking interface. `<Generated_SVG>` should appear as one continuous and complete vase. The outer wall, inner wall, top rim, and base should visually connect naturally without separated parts or assembly seams. Since Joint is None, screws, hinges, snap features, glue pads, slots, and protrusions should not appear. Minor triangular mesh texture or shading transitions caused by curved-surface rendering are acceptable.

    * **0 Points (Fail)**: The connection semantics contradict the task. For example, the vase is divided into multiple pieces that would need bonding or insertion; the top rim is detached from the body; the inner wall floats like a separate cup liner; the base is separated from the outer shell; or non-required joint features such as hinges, snap-fits, screw holes, or slots appear.

---

### 3. Tolerance

* **Evaluation Rubric (0-1 Score)**:

    * **1 Point (Pass)**: The manufacturing process is FDM Printing, treated under the 3D Printing precision range of 0.05-0.50 mm. Because this is not an assembly model, the tolerance check should focus on printable wall thickness, top rim thickness, the spacing between inner and outer walls, and bottom closure thickness rather than part-to-part fitting. Ask the judge to use the visible wall thickness at the top opening and the solid base appearance in `<Reference_SVG>` as the visual scale reference. `<Generated_SVG>` should show a clear hollow opening, an identifiable thick-wall structure, and a continuous rim. The wall should not be nearly zero-thickness, and it should not be so thick that the opening is almost blocked. Slight layer lines, mesh artifacts, or surface discretization are acceptable for FDM-style output.

    * **0 Points (Fail)**: The wall-thickness or opening tolerance is severely wrong. For example, the top opening disappears and the object becomes a solid sculpture; the inner and outer wall boundaries are visually chaotic and the cavity cannot be identified; the wall is paper-thin and likely to break under FDM printing; the wall thickness is so exaggerated that the opening becomes unusably narrow or the body becomes bulky; or the bottom is not closed, making the object invalid as a container or as a slicable closed body.

---

### 4. Functional Adaptation

* **Evaluation Rubric (0-1 Score)**:

    * **1 Point (Pass)**: The must-have function, weighted around 70

    * **0 Points (Fail)**: The must-have function is completely lost. For example, the model is a solid block with no opening; there is no flat base or the bottom is pointed so it cannot stand; the top is sealed and cannot hold dried flowers; the form collapses, self-intersects, or breaks so badly that it is no longer recognizable as a vase; or the waves are so extreme that the body is torn, punctured, or twisted into an unusable abstract sculpture.

---

### 5. Usage Stability

* **Evaluation Rubric (0-1 Score)**:

    * **1 Point (Pass)**: Ask the judge to evaluate desktop standing stability, base support area, center-of-mass position, and safe wall thickness. The vase is non-load-bearing but should support its own weight and a small amount of lightweight dried flowers. The intended force path is: vase self-weight / dried flowers  thick shell and base  tabletop. `<Generated_SVG>` should resemble `<Reference_SVG>` by having a solid flat or near-flat base with sufficient contact area, and the bodys center of mass should remain roughly near the central vertical axis. The wave pattern should be distributed evenly enough that the vase does not visibly lean to one side. The wall thickness should appear sufficient for FDM-printed shell stiffness, and the top rim should not be too thin or broken.

    * **0 Points (Fail)**: The object would inevitably tip or fail structurally. For example, the bottom is a sharp point or has an extremely small contact area; the body bulges heavily to one side so the center of mass falls outside the base; the vase is extremely tall and narrow with an insufficient base; the wall is paper-thin and the rim is broken; or the base is missing or perforated so it cannot stand stably.

---

### 6. Manufacturability

* **Evaluation Rubric (0-1 Score)**:

    * **1 Point (Pass)**: The material is PLA and the manufacturing process is FDM Printing. PLA is easy to print but relatively low-strength, while FDM / 3D Printing supports complex geometry but requires a slicable closed manifold solid. Very thin walls, non-manifold edges, self-intersecting surfaces, and uncontrolled overhangs may cause print failure. `<Generated_SVG>` should show a single thick-wall body compatible with PLA/FDM printing: a continuous outer shell, a clear hollow interior, a closed base, a continuous top rim, and no obvious non-manifold cracks. The twisted waves may be complex, but they should remain smooth, continuous, non-self-intersecting, and free of unexplained floating surfaces or internal fragments.

    * **0 Points (Fail)**: `<Generated_SVG>` violates FDM / PLA manufacturing constraints. For example, it is not a single closed manifold body; the outer shell has large cracks or self-intersections; isolated surfaces float inside the cavity; the bottom is not closed; the wall is nearly zero-thickness; the top rim breaks into multiple independent pieces; or the design contains extremely thin, long-span suspended PLA structures that would collapse or snap during printing.

---

Now wait for the user to provide `<Task_Doc>`, `<Reference_Code>`, and `<Reference_SVG>`, then dynamically generate the task-specific rubric.
"""

\end{promptbox}

%% file: chapter/ref.bib
@article{pointercad,
  title={Pointer-{CAD}: unifying {B-Rep} and command sequences via pointer-based edges \& faces selection},
  author={Qi, Dacheng and Wang, Chenyu and Xu, Jingwei and Chu, Tianzhe and Zhao, Zibo and Liu, Wen and Ding, Wenrui and Ma, Yi and Gao, Shenghua},
  journal={arXiv preprint arXiv:2603.04337},
  year={2026}
}

@inproceedings{preintner2025evocad,
  title={{EvoCAD}: evolutionary {CAD} code generation with vision language models},
  author={Preintner, Tobias and Yuan, Weixuan and K{\"o}nig, Adrian and B{\"a}ck, Thomas and Raponi, Elena and Van Stein, Niki},
  booktitle={2025 IEEE 37th International Conference on Tools with Artificial Intelligence (ICTAI)},
  pages={504--511},
  year={2025},
  organization={IEEE}
}

@article{articad,
  title={{ArtiCAD}: articulated {CAD} assembly design via multi-agent code generation},
  author={Shui, Yuan and Guan, Yandong and Zhang, Zhanwei and Hu, Juncheng and Zhang, Jing and Xu, Dong and Yu, Qian},
  journal={arXiv preprint arXiv:2604.10992},
  year={2026}
}

@article{text2cad,
  title={{Text2CAD}: generating sequential cad designs from beginner-to-expert level text prompts},
  author={Khan, Mohammad S and Sinha, Sankalp and Sheikh, Talha U and Stricker, Didier and Ali, Sk A and Afzal, Muhammad Z},
  journal={Advances in Neural Information Processing Systems},
  volume={37},
  pages={7552--7579},
  year={2024}
}

@article{mallis2026cadtests,
  title   = {Text-to-CAD Evaluation with CADTests},
  author  = {Mallis, Dimitrios and Wang, Marco and Karadeniz, Ahmet Serdar and Ricci, Elisa and Kacem, Anis and Aouada, Djamila},
  journal = {arXiv preprint arXiv:2605.07807},
  year    = {2026}
}

@inproceedings{khan2024text2cad,
  title     = {Text2CAD: Generating Sequential {CAD} Designs from Beginner-to-Expert Level Text Prompts},
  author    = {Khan, Mohammad Sadil and Sinha, Sankalp and Sheikh, Talha Uddin and Stricker, Didier and Ali, Sk Aziz and Afzal, Muhammad Zeshan},
  booktitle = {Advances in Neural Information Processing Systems},
  volume    = {37},
  pages     = {7552--7579},
  publisher = {Curran Associates, Inc.},
  year      = {2024}
}

@article{wang2026text2cadbench,
  title={Text2CAD-Bench: A Benchmark for LLM-based Text-to-Parametric CAD Generation},
  author={Wang, Liang and Meng, Heng and Xiang, Zekai and Liu, Jin and Zhou, Pingyi and Chen, Litao and Tang, Yongqiang},
  journal={arXiv preprint arXiv:2605.18430},
  year={2026}
}

@article{mallis2026cadtestbench,
  title={Text-to-CAD Evaluation with CADTests},
  author={Mallis, Dimitrios and Wang, Marco and Karadeniz, Ahmet Serdar and Ricci, Elisa and Kacem, Anis and Aouada, Djamila},
  journal={arXiv preprint arXiv:2605.07807},
  year={2026}
}

@article{cadsmith_2026,
  title={CADSmith: Multi-Agent CAD Generation with Programmatic Geometric Validation},
  author={Barkley, Jesse and Loghmani, Rumi and Farimani, Amir Barati},
  journal={arXiv preprint arXiv:2603.26512},
  year={2026}
}

@InProceedings{Wu_2021_ICCV,
    author    = {Wu, Rundi and Xiao, Chang and Zheng, Changxi},
    title     = {{DeepCAD}: A deep generative network for computer-aided design models},
    booktitle = {Proceedings of the IEEE/CVF International Conference on Computer Vision (ICCV)},
    month     = {October},
    year      = {2021},
    pages     = {6772-6782}
}

@inproceedings{Khan_2024_Text2CAD,
    author = {Khan, Mohammad Sadil and Sinha, Sankalp and Sheikh, Talha Uddin and Stricker, Didier and Ali, Sk Aziz and Afzal, Muhammad Zeshan},
    title = {{Text2CAD}: generating sequential {CAD} models from beginner-to-expert level text prompts},
    booktitle = {Advances in Neural Information Processing Systems (NeurIPS)},
    year = {2024},
    pages = {7552--7579}
}

@inproceedings{Wang_2025_CADFusion,
    author = {Wang, Ruiyu and Yuan, Yu and Sun, Shizhao and Bian, Jiang},
    title = {{Text-to-CAD} generation through infusing visual feedback in large language models},
    booktitle = {Proceedings of the International Conference on Machine Learning (ICML)},
    year = {2025}
}

@inproceedings{Li_2024_CADTranslator,
    author = {Li, Xueyang and Song, Yu and Lou, Yunzhong and Zhou, Xiangdong},
    title = {{CAD} translator: an effective drive for text to {3D} parametric computer-aided design generative modeling},
    booktitle = {Proceedings of the ACM International Conference on Multimedia (ACM MM 2024), Poster},
    year = {2024}
}

@inproceedings{liao2025automated,
  title={Automated {CAD} modeling sequence generation from text descriptions via transformer-based large language models},
  author={Liao, Jianxing and Xu, Junyan and Sun, Yatao and Tang, Maowen and He, Sicheng and Liao, Jingxian and Yu, Shui and Li, Yun and Guan, Xiaohong},
  booktitle={Proceedings of the 63rd Annual Meeting of the Association for Computational Linguistics (Volume 1: Long Papers)},
  pages={21720--21748},
  year={2025}
}

@article{Yavartanoo_2024_Text2CAD,
    author = {Yavartanoo, Mohsen and Hong, Sangmin and Neshatavar, Reyhaneh and Lee, Kyoung Mu},
    title = {{Text2CAD}: text to {3D} {CAD} generation via technical drawings},
    journal = {arXiv preprint arXiv:2411.06206},
    year = {2024}
}

@article{alrashedy2024generating,
  title={Generating {CAD} code with vision-language models for {3D} designs},
  author={Alrashedy, Kamel and Tambwekar, Pradyumna and Zaidi, Zulfiqar and Langwasser, Megan and Xu, Wei and Gombolay, Matthew},
  journal={arXiv preprint arXiv:2410.05340},
  year={2024}
}

@article{xu2024cad,
  title={{CAD-MLLM}: Unifying multimodality-conditioned {CAD} generation with {MLLM}},
  author={Xu, Jingwei and Wang, Chenyu and Zhao, Zibo and Liu, Wen and Ma, Yi and Gao, Shenghua},
  journal={arXiv preprint arXiv:2411.04954},
  year={2024}
}
